\documentclass{article}
\usepackage[utf8]{inputenc}
\usepackage{float}
\usepackage[T1]{fontenc}
\usepackage{hyperref}
\usepackage{url}
\usepackage{booktabs}
\usepackage{amsfonts}
\usepackage{amsmath}
\usepackage{amssymb}
\usepackage{nicefrac}
\usepackage{microtype}
\usepackage{xcolor}
\usepackage{graphicx}
\usepackage{wrapfig}
\usepackage{textgreek}
\usepackage{float}
\usepackage{microtype}
\usepackage{natbib}
\usepackage{placeins}
\usepackage[margin=1in]{geometry}
\title{Adaptive Memory Decay for Log-Linear Attention}
\usepackage{authblk}

\author[1,*]{Yaxita Amin}
\author[1]{Helen Zichen Li}
\author[1]{Mengfan Zhang}
\author[1]{Samet Ayhan}

\affil[1]{University of Maryland}
\affil[*]{Corresponding author: yaxita@umd.edu}
\date{}
\begin{document}
\maketitle
\begin{abstract}
Sequence models face a fundamental tradeoff between memory capacity and computational efficiency. Transformers achieve expressive context modeling at quadratic cost, while linear attention and state-space models run in linear time by compressing context into a fixed-size hidden state, inherently limiting recall. Log-linear attention navigates this tradeoff by organizing memory across a Fenwick tree hierarchy, growing its hidden state logarithmically with sequence length at log-linear compute cost. However, its memory decay parameter $\lambda$ is fixed and independent of the input, assigning uniform weights across all hierarchy levels regardless of the content, which introduces unnecessary rigidity. We propose learning λ directly from the input via a lightweight two-layer MLP, producing per-token, per-level decay that adapts to content rather than position. A softplus activation lets each Fenwick tree level scale independently, avoiding the inter-level competition that softmax introduces. This modification preserves log-linear complexity exactly and adds negligible parameter overhead. We evaluate on associative recall, selective copying, and language modeling, finding that input-dependent decay consistently outperforms the baseline, with the largest gains in long-range memory settings where baseline λ degrades or collapses entirely.
\end{abstract}

\section{Introduction}
The Transformer architecture \citep{vaswani2017attention} is a core algorithm 
for sequence modeling, achieving rich context modeling by allowing every token 
in a sequence to attend to every other token. This enables high recall and 
long-range reasoning. However, with longer sequences, computational cost grows 
quadratically and memory grows linearly. Hardware-optimized kernels like 
FlashAttention \citep{dao2022flashattention,dao2023flashattention2} cut down 
runtime considerably, yet the fundamental quadratic scaling remains inescapable, 
making long-sequence modeling prohibitively expensive.

Linear transformers \citep{katharopoulos2020transformers} and state-space based 
models \citep{gu2021s4, gu2023mamba, dao2024mamba2} address this by compressing 
all past context into a fixed-size recurrent state, reducing training complexity 
to $\mathcal{O}(T)$ and decoding memory to $\mathcal{O}(1)$. Modern variants 
with data-dependent gating \citep{yang2023gla, sun2023retnet, peng2023rwkv} 
have significantly narrowed the quality gap with Transformers. However, the 
fixed-size hidden state remains a hard bottleneck: no matter how the gating is 
tuned, the model is fundamentally limited in how much information it can retain 
from arbitrarily long sequences, leading to measurable degradation on 
recall-intensive tasks \citep{arora2023zoology, arora2024simple}.

Log-Linear Attention \citep{guo2026loglinear} proposed a middle-ground solution 
by replacing the single recurrent state with a hierarchy of states organized via 
a Fenwick tree decomposition. Recent tokens are retained at fine granularity 
while older tokens are summarized at progressively coarser scales, with the 
total number of states growing only logarithmically with sequence length. This 
yields $\mathcal{O}(T \log T)$ training compute and $\mathcal{O}(\log T)$ 
decoding memory, strictly between linear and quadratic attention. The output at 
each timestep is a weighted sum across memory levels, controlled by a coefficient 
$\lambda_t^{(\ell)}$ per token per level. In the original formulation, this 
coefficient is computed as a near-fixed linear projection of the input hidden 
state, making it largely input-independent in practice. The original authors 
explicitly acknowledge this as an open limitation, noting that richer 
parameterization of $\lambda$ was promising but unexplored due to compute 
constraints \citep{guo2026loglinear}.%
\footnote{\url{https://openreview.net/forum?id=mOJgZWkXKW} --- see author 
response and reviewer discussion regarding $\lambda$ parameterization.} 
This is further confirmed in the OpenReview discussion, where the authors 
accepted adaptive $\lambda$ parameterization as a limitation and explicitly 
identified it as a promising direction for future work.

We take this as our starting point. Deciding how much weight to assign each 
memory level is an inherently nonlinear, content-dependent problem: a token 
closing a long-range dependency should draw heavily on coarse distant summaries, 
while a token responding to its immediate context should emphasize fine-grained 
recent states. A linear projection cannot flexibly express this. We propose 
replacing it with a lightweight two-layer MLP with softplus activation, 
producing a distinct $\lambda_t^{(\ell)}$ for every token and every hierarchy 
level. Softplus is chosen specifically to allow each level to scale 
independently, in contrast to softmax, which forces competition between levels 
and artificially suppresses some when others are active. This modification adds 
negligible parameters and preserves $\mathcal{O}(T \log T)$ complexity exactly. 
We summarize our contributions as follows:

\begin{itemize}
    \item \textbf{Content-adaptive memory weighting:} The linear $\lambda$ 
    projection in log-linear attention is replaced with a lightweight two-layer 
    MLP, producing per-token, per-level memory decay that adapts to content 
    rather than position.

    \item \textbf{Softplus as the principled activation:} Softplus enables 
    independent per-level scaling, in contrast to softmax, which introduces 
    harmful inter-level competition between memory hierarchy levels.

    \item \textbf{Complexity-preserving design:} The modification preserves 
    $\mathcal{O}(T \log T)$ training complexity and $\mathcal{O}(\log T)$ 
    decoding memory exactly, adding less than $0.007\%$ parameter overhead.

    \item \textbf{Empirical validation:} Consistent improvements over the 
    baseline are observed on multi-query associative recall, selective copying, 
    and WikiText-103 language modeling, with the largest gains at high memory 
    loads and long sequences where the original parameterization degrades or 
    collapses.
\end{itemize}
\label{sec:intro}

\section{Background}
\label{sec:background}

\subsection{Softmax Attention}


Softmax attention is a central component of the Transformer architecture \citep{vaswani2017attention}. Given query, key, and value projections $Q,K,V \in \mathbb{R}^{n \times d}$, scaled dot-product attention is defined as
\begin{equation}
\mathrm{Attention}(Q,K,V) =
\mathrm{softmax}\left(\frac{QK^\top}{\sqrt{d}}\right)V
\end{equation}
This mechanism allows each token to directly interact with every other token in the sequence, making it highly effective for modeling long-range dependencies.

The main limitation is that the attention matrix $QK^\top$ has size $n \times n$, giving softmax attention $\mathcal{O}(n^2)$ time and memory complexity with respect to sequence length. FlashAttention and FlashAttention-2 reduce the practical cost of exact attention by avoiding materialization of the full attention matrix and improving GPU utilization \citep{dao2022flashattention,dao2023flashattention2}. However, these methods still preserve the underlying quadratic dependence on sequence length, motivating alternatives that change the memory structure of attention itself.
\subsection{Linear Attention and State Space Models}


A major line of work addresses the quadratic cost of softmax attention by replacing explicit pairwise token interactions with recurrent or state-based computation. Linear attention uses kernel feature maps to rearrange the attention computation so that the full $n \times n$ matrix does not need to be explicitly constructed \citep{katharopoulos2020transformers}. In the causal setting, this gives attention a recurrent interpretation: instead of storing all previous keys and values separately, the model maintains a running hidden state that summarizes past context.

State space models follow a similar motivation from a dynamical systems perspective. The S4 model showed that structured state space parameterizations can make recurrent sequence models practical for long-range modeling \citep{gu2021s4}. Later models add stronger memory control: Mamba uses input-dependent selective state updates \citep{gu2023mamba}, Gated Linear Attention adds data-dependent gates \citep{yang2023gla}, and RetNet and RWKV combine recurrent inference with Transformer-like parallel training \citep{sun2023retnet,peng2023rwkv}. These models improve efficiency, but they still compress past context into one or a small number of hidden states. This fixed-state compression can create a recall bottleneck when a model must recover precise information from earlier in the sequence, motivating memory structures that remain efficient while preserving richer access to past context.
\subsection{Log-Linear Attention}
\begin{wrapfigure}{r}{0.3\textwidth}
  \centering
  \vspace{-8pt}
  \includegraphics[width=0.3\textwidth]{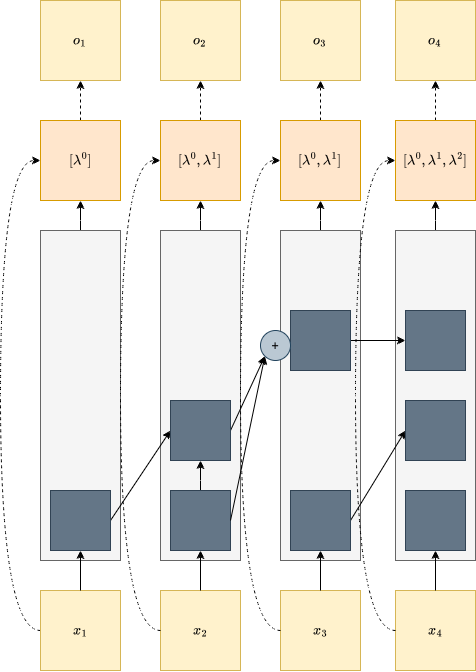}
  \caption{Fenwick-tree memory structure in log-linear attention. At each
  timestep $t$, the prefix is partitioned into hierarchical memory buckets
  $M_t^{(\ell)}$ at increasing temporal scales. The output $o_t$ is a
  weighted sum across all active levels, controlled by $\lambda_t^{(\ell)}$.}
  \label{fig:fenwick}
  \vspace{-8pt}
\end{wrapfigure}
Log-Linear Attention addresses the fixed-state limitation of linear attention 
and state space models by replacing a single recurrent memory with a 
logarithmically growing hierarchy of hidden states \citep{guo2026loglinear}. 
Instead of storing every past token as in full attention or compressing the 
entire prefix into one state as in linear attention, log-linear attention 
partitions the prefix into memory buckets at different temporal scales. This 
hierarchy is implemented through a Fenwick-tree decomposition, giving at most 
$L=\mathcal{O}(\log T)$ memory levels. Lower levels preserve recent context 
at finer resolution, while higher levels summarize older context more coarsely.

A key part of this mechanism is the memory-level weighting term $\lambda$, 
which determines how much each level contributes to the output at a given 
timestep. This parameterization is important because different tokens may 
require different memory strategies: for example, a query token may need to 
retrieve information from a specific past segment, while a filler or padding 
token may not need the same memory access pattern. If $\lambda$ is too static 
or only weakly input-dependent, the model may apply nearly the same memory 
strategy across very different tokens.

Guo et al. identify the parameterization of $\lambda$ as an underexplored 
design choice \citep{guo2026loglinear}. Our work builds on this point by 
replacing the original $\lambda$ computation with an MLP-parameterized 
version, allowing memory-level weights to depend more strongly on the current 
input while preserving the same log-linear memory structure.

Log-linear attention \citep{guo2026loglinear} organizes past context
into a Fenwick-tree hierarchy of hidden states. At each timestep $t$,
the sequence prefix is partitioned into at most $L = \lceil \log_2 T
\rceil$ memory levels, where level $\ell$ summarizes a power-of-two
span of past tokens at a coarser temporal resolution as $\ell$
increases. The output at timestep $t$ is a weighted combination across
all active memory levels:
\begin{equation}
  o_t = \sum_{\ell=0}^{L_t - 1} \lambda_t^{(\ell)} \cdot M_t^{(\ell)},
  \label{eq:output}
\end{equation}
where $M_t^{(\ell)}$ is the hidden state at level $\ell$ and
$\lambda_t^{(\ell)}$ controls its contribution. In the original
formulation, $\lambda_t^{(\ell)}$ is computed as a linear projection
of the input followed by a fixed activation, producing weights that
are only weakly dependent on the input content. As the original authors note in both the paper's limitations section and the 
OpenReview discussion, this parameterization was left underexplored due to 
compute constraints \citep{guo2026loglinear}.
\section{Method}
\label{sec:method}

\subsection{Motivation}
\label{sec:motivation}

The memory-level weighting $\lambda_t^{(\ell)}$ plays a critical role
in determining which temporal scales the model attends to at each
step. A token that closes a long-range dependency should draw heavily
on coarse, distant memory levels, while a token responding to
immediate context should weight fine-grained recent levels. This is an
inherently content-dependent, nonlinear decision that a linear
projection cannot flexibly express.

Importantly, $\lambda$ should respond to \textit{content} rather than 
\textit{position}. The original authors explored a RoPE-based 
parameterization in early experiments,
\begin{equation}
    \lambda_t^{(\ell)} = \mathrm{softplus}\!\left(\tilde{\lambda}_t^{(\ell)}\right), 
    \quad 
    \tilde{\lambda}_t^{(\ell)} = \left(\mathrm{rope}(\mathbf{x}_t, \mathbf{W}_0, \ell) 
    \mathbf{W}_1\right) \odot \left(\mathrm{rope}(\lambda, \ell)\, \mathbf{W}_1'\right),
\end{equation}
but did not adopt it in the main work due to higher computational cost 
(OpenReview, official author response, Nov 2025).\footnote{\url{https://openreview.net/forum?id=mOJgZWkXKW}} 
Crucially, RoPE encodes \textit{position} rather than \textit{content}, 
which is the wrong inductive bias for $\lambda$ --- memory-level weighting 
should depend on what a token is, not where it appears. Our MLP 
parameterization avoids this by operating directly on the input hidden 
state, producing content-adaptive weights at lower computational cost 
than the RoPE variant.

Moreover, the choice of activation matters. Softmax normalizes weights
across levels so that emphasizing one level suppresses all others,
introducing artificial competition between memory scales that may
prevent the model from simultaneously drawing on both recent and
distant context. Softplus imposes no such constraint: each level is
activated independently, allowing the model to freely combine multiple
temporal scales when the input calls for it.

\subsection{LambdaMLPSoftplus}
\setlength{\footnotesep}{0pt}
\label{sec:method_mlp}

\begin{wrapfigure}{r}{0.40\textwidth}
\centering
  \vspace{-8pt}
  \includegraphics[width=0.34\textwidth, keepaspectratio]{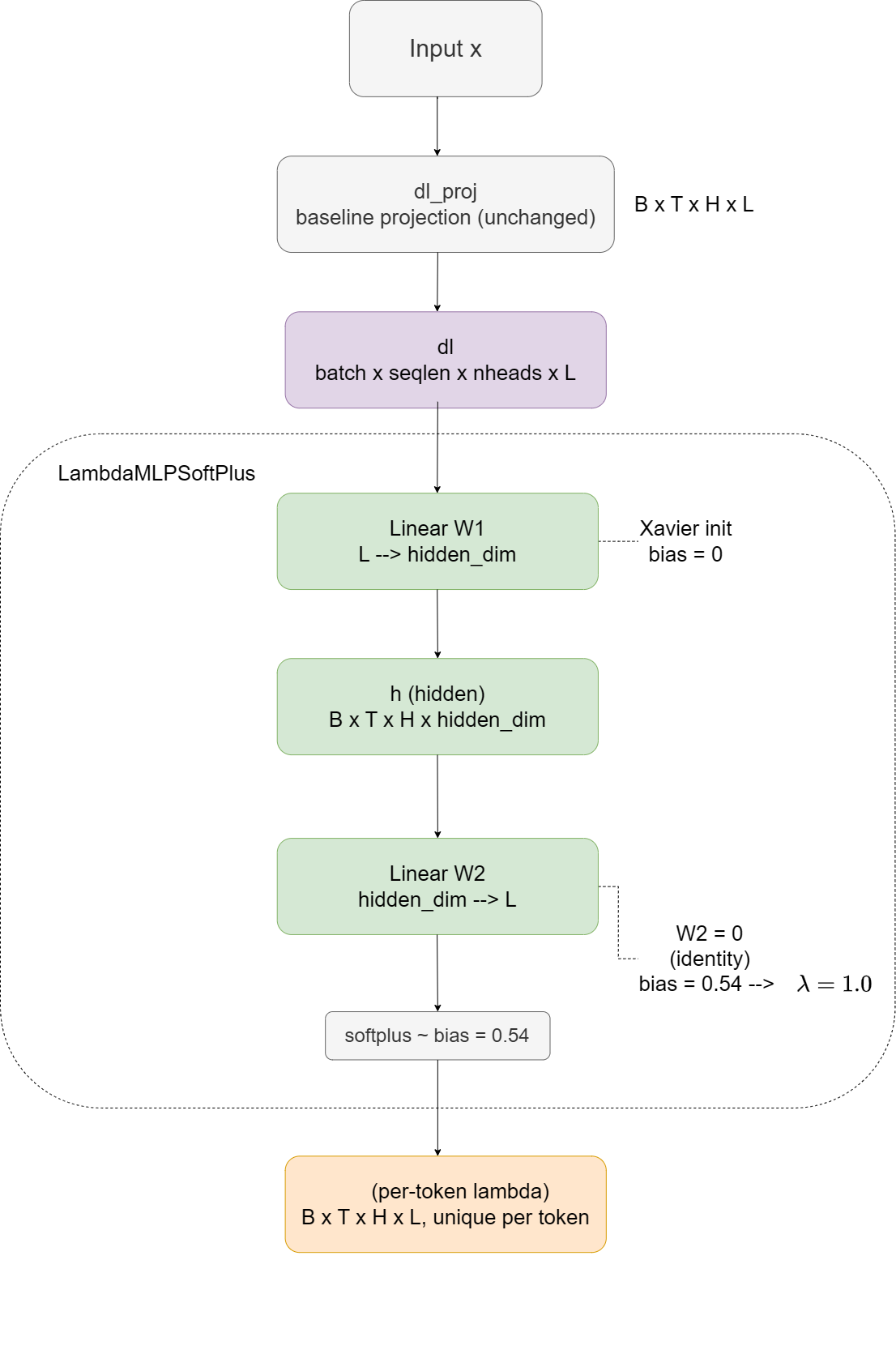}
  \caption{Architecture of LambdaMLPSoftplus. The baseline projection
  $\mathbf{d}_t$ is passed through two linear layers with softplus activation.
  Initialization ensures $\lambda \approx 1.0$ at the start of training.}
  \label{fig:arch}
  \vspace{-8pt}
\end{wrapfigure}

We replace the baseline $\lambda$ projection with a lightweight two-layer MLP. 
Concretely, let $\mathbf{d}_t \in \mathbb{R}^{H \times L}$ be the intermediate 
representation produced by the baseline projection for token $t$ across all 
heads $H$ and levels $L$, which we keep unchanged. Our MLP takes this as input 
and produces content-adaptive weights:
\begin{equation}
  \mathbf{h}_t = \mathbf{d}_t \mathbf{W}_1, \qquad
  \lambda_t = \mathrm{softplus}\!\left(\mathbf{h}_t \mathbf{W}_2 + b\right),
  \label{eq:mlp}
\end{equation}
where $\mathbf{W}_1 \in \mathbb{R}^{L \times d_h}$, $\mathbf{W}_2 \in 
\mathbb{R}^{d_h \times L}$, and $b$ is a scalar bias. The output $\lambda_t 
\in \mathbb{R}^{B \times T \times H \times L}$ provides a unique per-token, 
per-head, per-level weight.

\paragraph{Initialization.} $\mathbf{W}_1$ is initialized with Xavier uniform 
initialization with bias zero. $\mathbf{W}_2$ is initialized to zero and the 
bias is set to $0.54$, so that $\mathrm{softplus}(0.54) \approx 1.0$ at 
initialization. This ensures the MLP starts as a near-identity transformation, 
matching the baseline $\lambda$ behavior at the start of training and allowing 
stable optimization from a well-defined starting point.

\paragraph{Complexity.} The MLP operates on the already-computed intermediate 
representation $\mathbf{d}_t$ of shape $B \times T \times H \times L$, where 
$L = \mathcal{O}(\log T)$. Both linear layers are over the $L$-dimensional 
level axis only, so the total added compute is $\mathcal{O}(T \cdot H \cdot L 
\cdot d_h)$, which is absorbed into the existing $\mathcal{O}(T \log T)$ term. 
Decoding memory is unchanged at $\mathcal{O}(\log T)$. With a hidden dimension 
$d_h = 2L$, the MLP adds fewer than $0.007\%$ additional parameters to the 
full model, making the overhead negligible in practice.

\subsection{Complexity Analysis}
\label{sec:complexity}

Table~\ref{tab:complexity} summarizes the computational properties of
the baseline and our method. Training complexity, decoding memory, and
the Fenwick-tree structure are all preserved exactly. The only change
is the parameterization of $\lambda$, which goes from a linear
projection to a two-layer MLP while remaining over the
$\mathcal{O}(\log T)$-dimensional level axis.

\begin{table}[h]
  \caption{Complexity comparison. Our modification preserves the
  log-linear profile exactly.}
  \label{tab:complexity}
  \centering
  \begin{tabular}{lccc}
    \toprule
    Method & Train & Decode memory & $\lambda$ params \\
    \midrule
    Softmax attention      & $\mathcal{O}(T^2)$      &
    $\mathcal{O}(T)$      & -- \\
    Linear attention       & $\mathcal{O}(T)$        &
    $\mathcal{O}(1)$      & -- \\
    Log-linear (baseline)  & $\mathcal{O}(T \log T)$ &
    $\mathcal{O}(\log T)$ & linear proj \\
    Log-linear (ours)      & $\mathcal{O}(T \log T)$ &
    $\mathcal{O}(\log T)$ & 2-layer MLP \\
    \bottomrule
  \end{tabular}
\end{table}

\section{Experiments}
\label{sec:experiments}

\subsection{Experimental Setup}
\label{sec:setup}

We evaluate three $\lambda$ parameterizations: the baseline fixed
projection, MLP-$\lambda$ with softplus activation
(LambdaMLPSoftplus), and MLP-$\lambda$ with softmax activation
(LambdaMLPSoftmax). All MLP variants use a two-layer architecture
with hidden dimension 64 and GELU activation, as described in
Section~\ref{sec:method_mlp}. All experiments use the Adam optimizer
with learning rate $10^{-3}$ and batch size 64. Models are trained
with 3--5 random seeds per configuration. Full hyperparameter details
are provided in Appendix~\ref{appendix:details}.We select $d_h = 64$ based on an ablation study (Appendix~\ref{appendix:dim_ablation}) 
showing it is the minimal sufficient size --- larger dimensions provide no consistent 
improvement on MQAR.

\subsection{Multi-Query Associative Recall (MQAR)}
\label{sec:mqar}

\paragraph{Task.} Multi-query associative recall \citep{arora2023zoology}
tests whether a model can retrieve values associated with keys presented
earlier in the sequence. Each sequence contains a set of key-value pairs
followed by queries; the model must output the correct value for each
query. We vary the number of key-value pairs $k \in \{4, 8, 16, 32\}$
at sequence length 128, training models with hidden dimension 64 for
5000 steps over 5 random seeds.

\begin{table}[h]
  \vspace{-8pt}
  \caption{MQAR accuracy (\%) at seq=128 across kv pairs.
  Mean and peak over 5 seeds. Baseline $\lambda$ shows high variance
  and collapses at harder settings.}
  \label{tab:mqar}
  \centering
  \small
  \begin{tabular}{llcccc}
    \toprule
    Mode & Metric & kv=4 & kv=8 & kv=16 & kv=32 \\
    \midrule
    Baseline $\lambda$       & Mean & 35.1 & 35.8 & 34.8 & 2.9  \\
                             & Peak & 99.4 & 99.5 & 99.3 & 99.3 \\
    MLP-$\lambda$ (softmax)  & Mean & 2.7  & 66.8 & 67.0 & 35.2 \\
                             & Peak & 3.8  & 99.3 & 99.6 & 99.0 \\
    MLP-$\lambda$ (softplus) & Mean & \textbf{67.0} & \textbf{66.7} & \textbf{66.7} & \textbf{57.9} \\
                             & Peak & \textbf{99.7} & 99.2 & 99.2 & \textbf{99.6} \\
    \bottomrule
  \end{tabular}
  \vspace{-5pt}
\end{table}
\begin{figure}[h]
  \centering
  \includegraphics[width=0.65\textwidth]{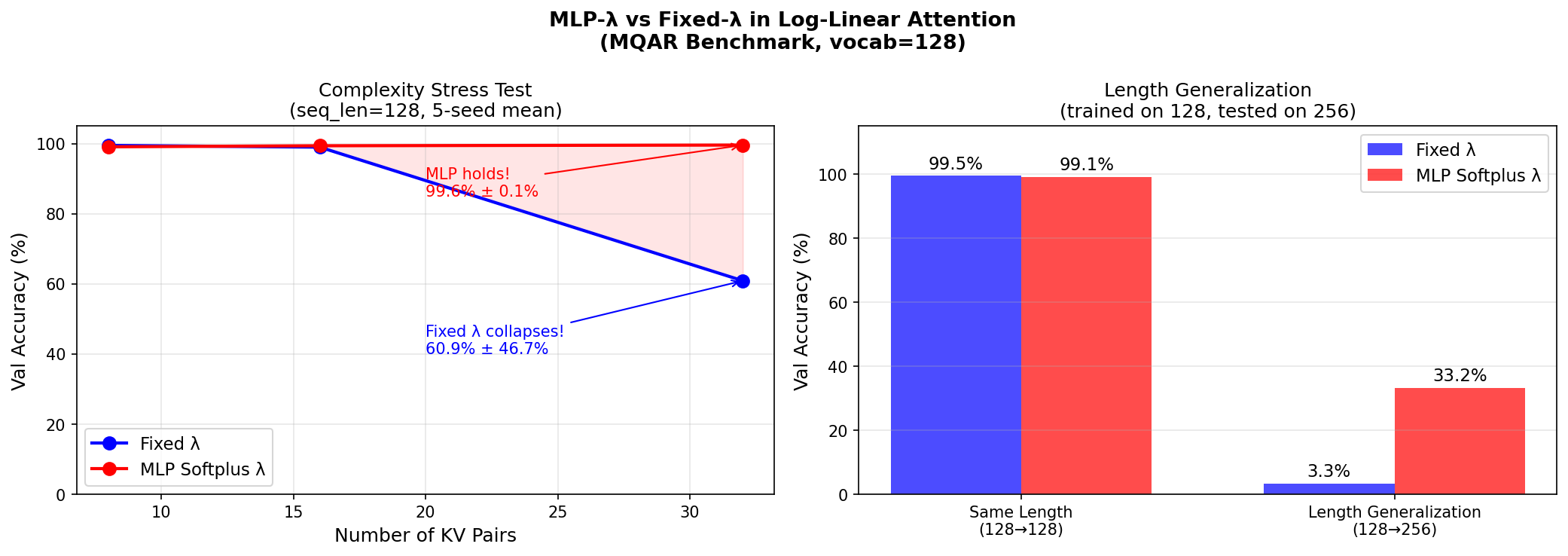}
  \caption{Complexity stress test (left) and length generalization (right). Softplus maintains accuracy as kv scales; retains 33.2\% at seq=256 vs baseline 3.3\%.}
  \label{fig:mqar}
\end{figure}

\paragraph{Results.} Table~\ref{tab:mqar} reports mean and peak
accuracy across seeds. Figure~\ref{fig:mqar} shows the complexity
stress test and length generalization results.

The key finding is that baseline $\lambda$ exhibits high variance
across seeds, either converging near-perfectly or collapsing entirely
to near-random accuracy. At kv=32, the baseline achieves only
$60.9\% \pm 46.7\%$ mean accuracy across 5 seeds, while
MLP-$\lambda$ (softplus) holds at $99.6\% \pm 0.1\%$. \footnote{Numbers reflect a separate run; Table~\ref{tab:mqar} reports a 
different random seed configuration.} while MLP-$\lambda$ (softplus) holds at 
$99.6\% \pm 0.1\%$.This instability
is consistent across longer sequences: trained on seq=128 and evaluated
on seq=256, the baseline drops to 3.3\% while softplus retains 33.2\%,
demonstrating stronger length generalization
(Figure~\ref{fig:mqar}). MLP-$\lambda$ (softmax) performs well at
moderate kv counts but fails at kv=32, consistent with the
inter-level competition hypothesis described in
Section~\ref{sec:motivation}.

\subsection{Selective Copying}
\label{sec:selective}
\paragraph{Task.} Selective copying requires the model to identify and
reproduce 16 specific tokens from a noisy sequence of length
$\in \{256, 512, 1024\}$, testing whether the model can selectively
retain relevant tokens while ignoring distractors. Models are trained
for 30000 steps with hidden dimension 64 and 5 seeds per configuration.
\begin{figure}[H]   
  \centering
  \includegraphics[width=0.46\textwidth, height=0.20\textheight, keepaspectratio]{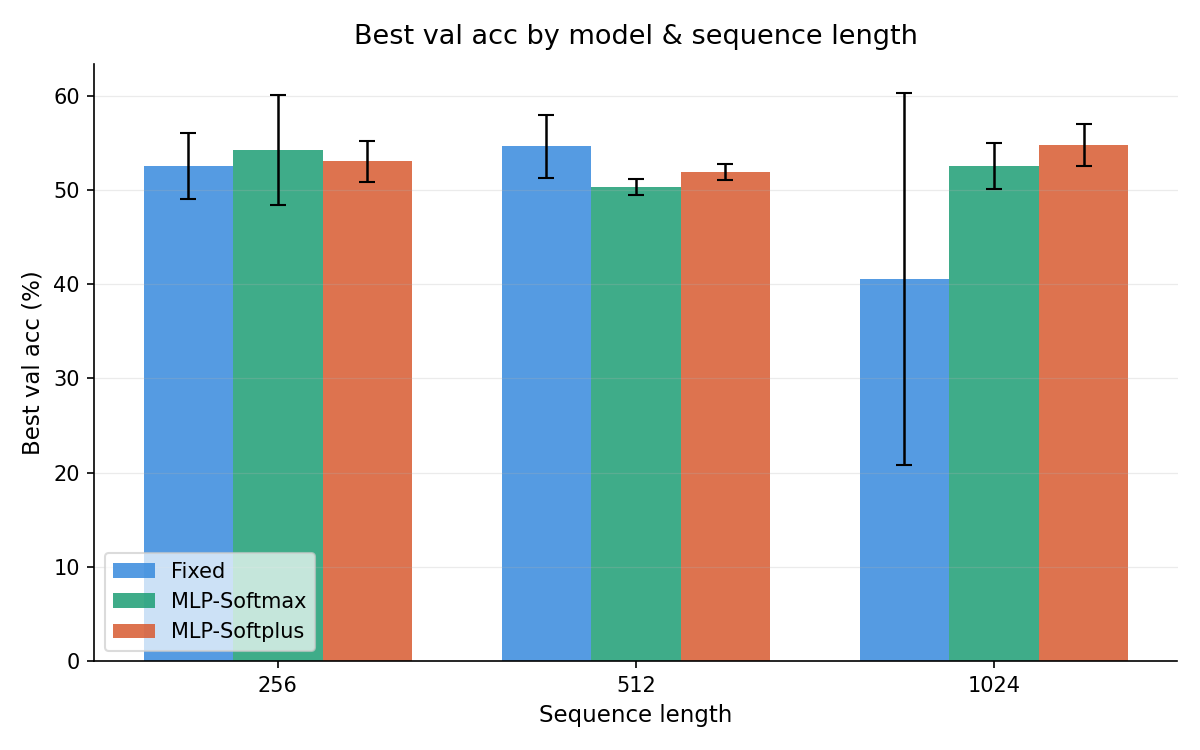}
  \includegraphics[width=0.46\textwidth, height=0.20\textheight, keepaspectratio]{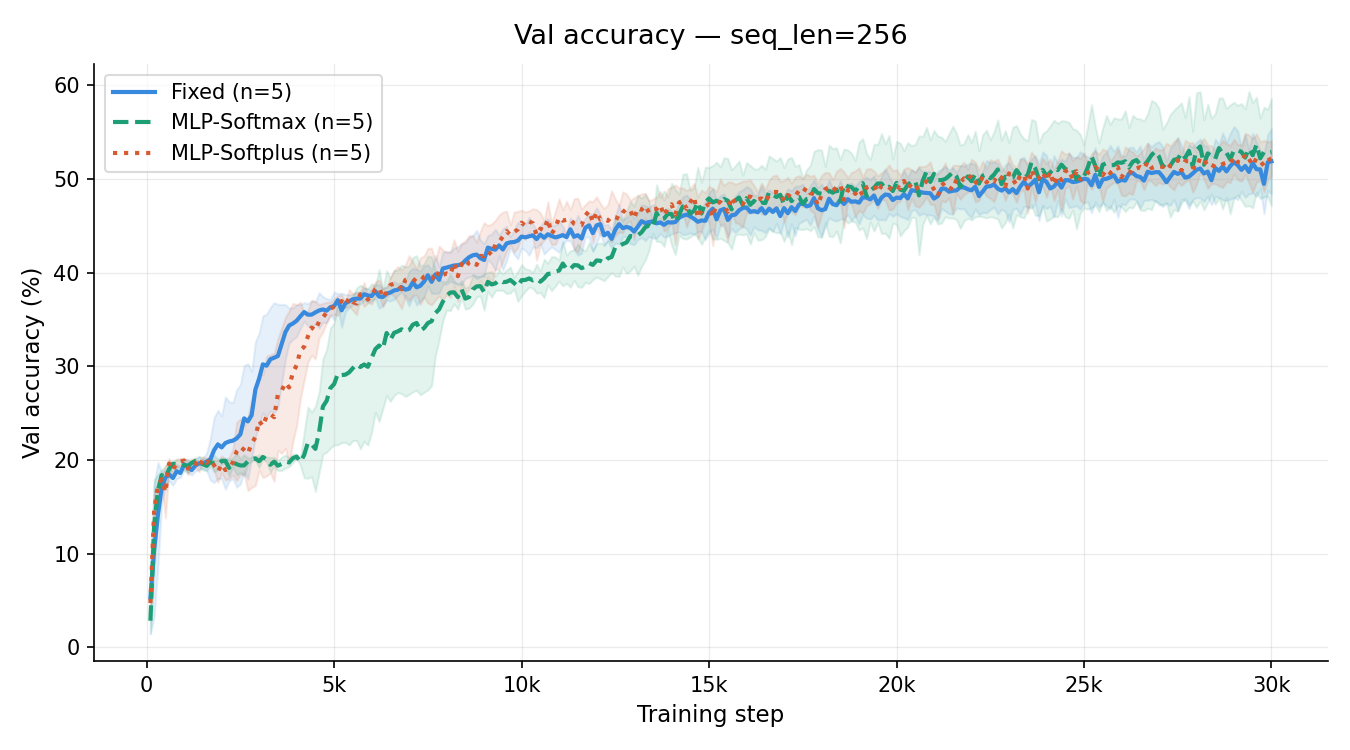}\\[2pt]
  \includegraphics[width=0.46\textwidth, height=0.20\textheight, keepaspectratio]{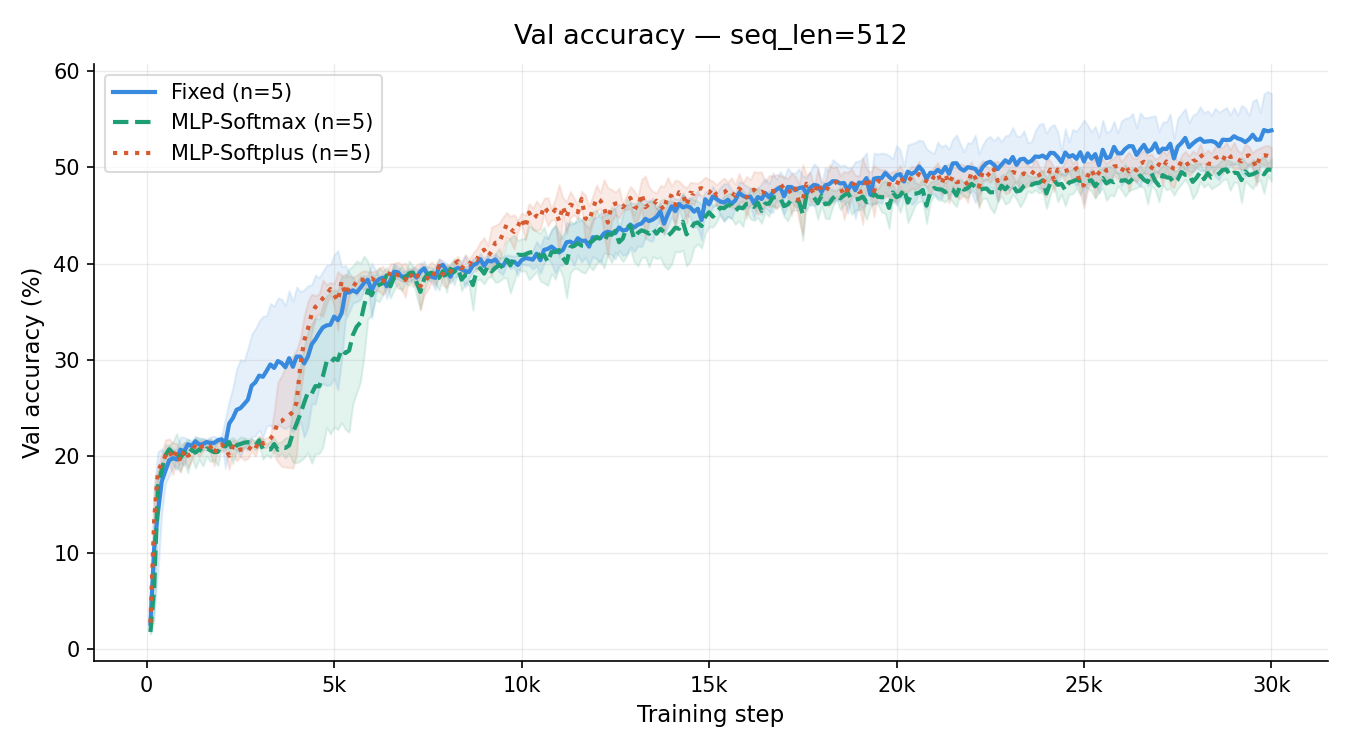}
  \includegraphics[width=0.46\textwidth, height=0.20\textheight, keepaspectratio]{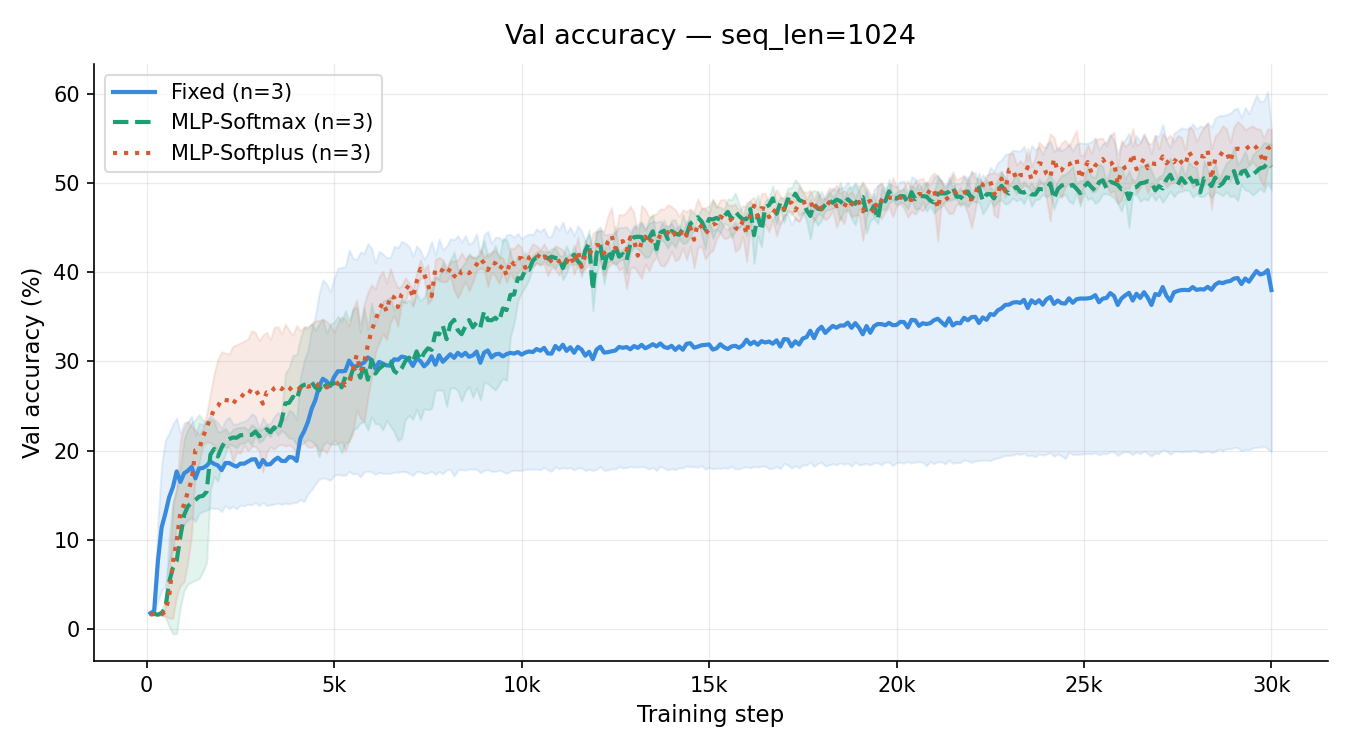}
  \caption{Selective copying results. Top-left: best validation accuracy
  across sequence lengths. Top-right and bottom: learning curves at
  seq=256, 512, and 1024. At seq=1024 the baseline plateaus near
  40\% while both MLP variants converge near 52--55\%.}
  \label{fig:selective}
\end{figure}
\paragraph{Results.}
\vspace{-0.5pt}
\begin{table}[H]
  \caption{Selective copying accuracy (\%) across sequence lengths.
  Mean $\pm$ std over 5 seeds (seq=256, 512) and 3 seeds (seq=1024).}
  \label{tab:selective}
  \centering
  \begin{tabular}{lccc}
    \toprule
    Mode & seq=256 & seq=512 & seq=1024 \\
    \midrule
    Baseline $\lambda$       & 52.6 $\pm$ 3.5 & 54.6 $\pm$ 3.4 & 40.6 $\pm$ 19.8 \\
    MLP-$\lambda$ (softplus) & 53.1 $\pm$ 2.2 & 51.9 $\pm$ 0.8 & \textbf{54.8} $\pm$ 2.3 \\
    MLP-$\lambda$ (softmax)  & \textbf{54.3} $\pm$ 5.8 & 50.3 $\pm$ 0.8 & 52.6 $\pm$ 2.5 \\
    \bottomrule
  \end{tabular}
\end{table}

Results are shown in Table~\ref{tab:selective} and
Figure~\ref{fig:selective}. At seq=256 and seq=512 performance is
broadly comparable across methods, with differences within the margin
of variance. The clearest signal emerges at seq=1024, where the
baseline degrades to $40.6 \pm 19.8\%$ --- a standard deviation nearly
five times larger than either MLP variant --- while MLP-$\lambda$
(softplus) achieves $54.8 \pm 2.3\%$, a gap of 14.2 points. This high
variance in the baseline reflects instability across seeds: the model
either partially learns the task or collapses entirely, consistent with
the optimization fragility observed in MQAR. Both MLP variants remain
stable across seeds ($\pm$2.3--2.5\%), confirming that adaptive memory
weighting not only improves mean accuracy but also substantially reduces
training instability under high memory pressure. The learning curves
confirm this divergence: both MLP variants climb steadily through 30k
steps while the baseline plateaus well below them.
\subsection{Language Modeling}
\label{sec:lm}
\paragraph{Task.} We evaluate on WikiText-103 \citep{merity2016wikitext}
using the GPT-2 tokenizer with vocabulary size 50,257. We report results
for two model configurations: a larger model with hidden size 512, 6
transformer layers, 4 attention heads, and head dimension 64; and a
smaller model with hidden size 256, 6 transformer layers, 4 attention
heads, and head dimension 64. Both use AdamW with learning rate
$3 \times 10^{-4}$, weight decay 0.1, linear warmup over 500 steps
followed by cosine decay, batch size 8, and sequence length 512 for
40,000 steps. Validation perplexity is reported on the WikiText-103
validation split.

\paragraph{Results.}
\begin{table}[h]
  \caption{Language modeling perplexity on WikiText-103 validation
  set. Lower is better. Results at 40k steps, seq length 512.
  Hidden size 512 results are single-seed; hidden size 256 results
  are mean $\pm$ std over 3 seeds.}
  \label{tab:lm}
  \centering
  \begin{tabular}{llcc}
    \toprule
    Mode & Hidden & Steps & Val PPL \\
    \midrule
    Baseline $\lambda$       & 512 & 40k & 224.71 \\
    MLP-$\lambda$ (softmax)  & 512 & 40k & 226.06 \\
    MLP-$\lambda$ (softplus) & 512 & 40k & \textbf{218.57} \\
    \midrule
    Baseline $\lambda$       & 256 & 40k & $257.02 \pm 0.29$ \\
    MLP-$\lambda$ (softmax)  & 256 & 40k & $257.31 \pm 0.25$ \\
    MLP-$\lambda$ (softplus) & 256 & 40k & $\mathbf{256.82 \pm 0.20}$ \\
    \bottomrule
  \end{tabular}
\end{table}

Table~\ref{tab:lm} reports results across two model sizes.
At hidden size 512, MLP-$\lambda$ (softplus) achieves 218.57
perplexity, outperforming baseline $\lambda$ by 6.14 points and
MLP-$\lambda$ (softmax) by 7.49 points. At hidden size 256, averaged
over 3 seeds, MLP-$\lambda$ (softplus) achieves $256.82 \pm 0.20$,
again outperforming both the baseline ($257.02 \pm 0.29$) and
MLP-$\lambda$ (softmax) ($257.31 \pm 0.25$). The gains are smaller
at the reduced model size, consistent with adaptive $\lambda$
providing more benefit when the model has greater representational
capacity. Across both configurations, softplus consistently
outperforms the baseline, suggesting that the benefit of adaptive
memory decay extends beyond synthetic recall tasks to natural
language modeling.

\subsection{Lambda Visualization}
\label{sec:lambda_viz}

\begin{figure}[h]
  \centering
  \includegraphics[width=0.49\textwidth]{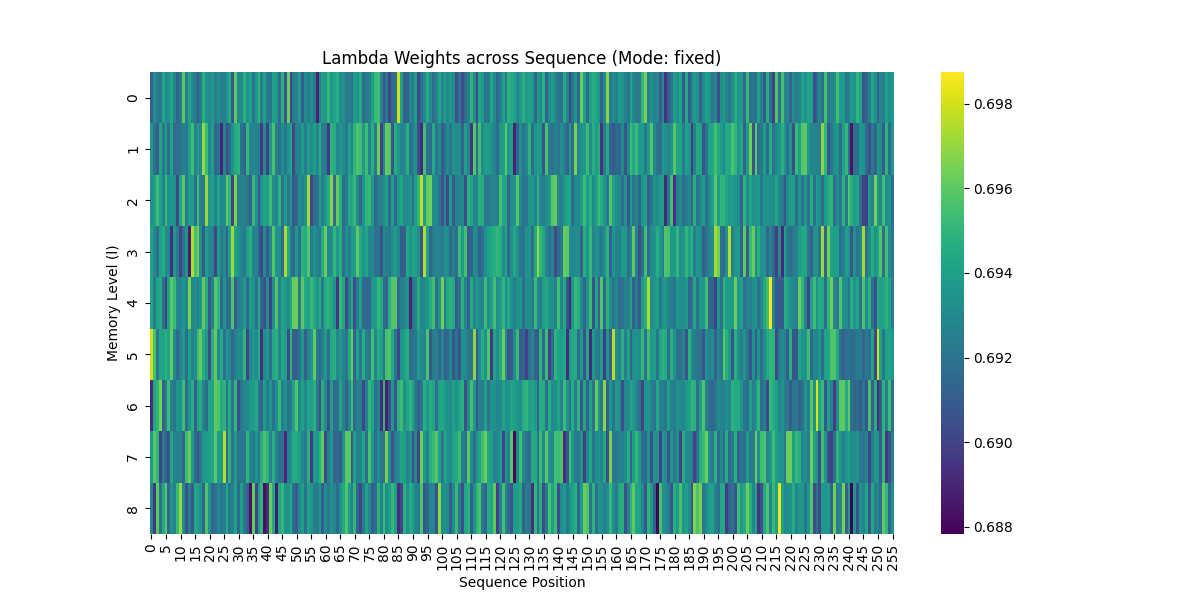}
  \includegraphics[width=0.49\textwidth]{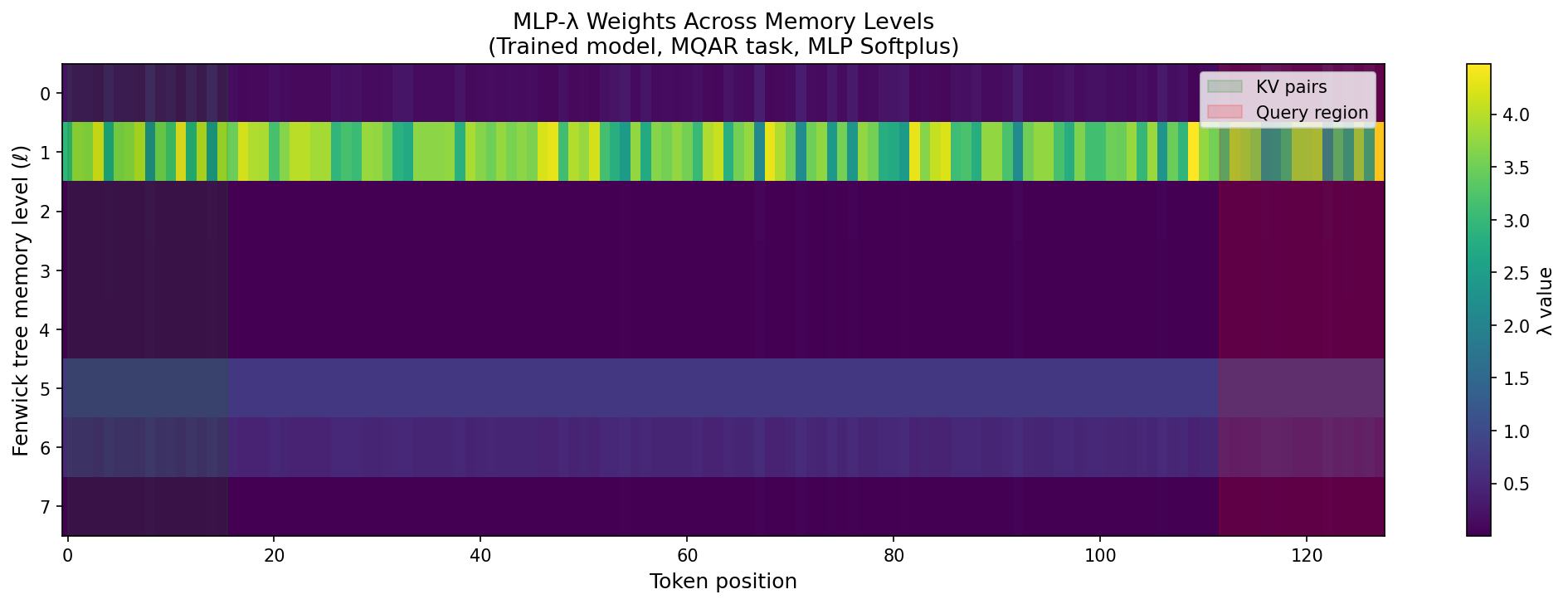}
  \caption{Left: baseline $\lambda$ weights show near-uniform values 
  across all memory levels and token positions (range 0.688--0.698), 
  confirming input-independence. Right: MLP-$\lambda$ (softplus) 
  learns sharp content-dependent patterns, with level 1 strongly 
  activated at key-value and query tokens.}
  \label{fig:lambda_viz}
\end{figure}

Figure~\ref{fig:lambda_viz} visualizes the per-token $\lambda$ weights
learned by MLP-$\lambda$ (softplus) on the MQAR task. In contrast to
baseline $\lambda$, which applies a nearly uniform weighting across all
hierarchy levels regardless of token content, the MLP-parameterized
version learns highly structured and content-dependent patterns. Level
1 receives strong activation during key-value pair tokens (green
regions), while query tokens (pink regions) trigger a sharp spike in
level 1 weights with near-zero activation at deeper levels. This
pattern suggests the model has learned to route memory access through
the fine-grained recent context level when resolving associative
queries, rather than distributing attention uniformly across the
Fenwick tree hierarchy.

\section{Related Work}
\label{sec:related}


Efficient attention mechanisms have been studied from both systems and algorithmic perspectives. FlashAttention and FlashAttention-2 improve exact softmax attention through I/O-aware tiling, kernel fusion, and better GPU parallelization, allowing attention to run faster without materializing the full attention matrix \citep{dao2022flashattention,dao2023flashattention2}. However, these methods retain the quadratic dependence on sequence length. A separate line of work replaces softmax attention with linear or recurrent formulations, where past tokens are summarized into compact states instead of stored through an explicit $n \times n$ attention matrix \citep{katharopoulos2020transformers}. Recent systems efforts such as Lightning Attention, Flash Linear Attention, and Flame make these efficient models practical through optimized kernels and training frameworks \citep{qin2024lightning,yang2024fla,zhang2025flame}. Our work is complementary to these efforts because we keep the log-linear computational structure fixed and study how its memory weighting mechanism can be made more adaptive.


Linear recurrent models and state space models provide another path toward efficient long-context modeling. S4 made structured state space models practical for long sequences \citep{gu2021s4}, while Mamba and Mamba-2 introduced more expressive selective state updates and connected SSMs to attention through state space duality \citep{gu2023mamba,dao2024mamba2}. Related models such as Gated Linear Attention, RetNet, DeltaNet, and Gated DeltaNet further explore how recurrent states should be updated, gated, and reused over time \citep{yang2023gla,sun2023retnet,sun2024deltanet,yang2024gateddelta}. Other linear RNN variants, including fast weight programmers, GateLoop, HGRN, HGRN2, and negative-eigenvalue linear RNNs, similarly show that memory dynamics strongly affect expressivity \citep{schlag2021linear,hao2023gateloop,qin2023hgrn,hao2023hgrn2,grazzi2024negative}. Our work is motivated by the same issue of memory control, but focuses on the memory-level weighting term in log-linear attention rather than changing the recurrent transition rule itself.


The RWKV family is another example of efficient sequence modeling through recurrent memory. RWKV combines Transformer-like parallel training with RNN-like constant-memory inference using a Receptance Weighted Key Value architecture \citep{peng2023rwkv}. Later variants such as Eagle, Finch, and RWKV-7 improve the recurrent state through matrix-valued states, dynamic recurrence, vector-valued gating, and more input-dependent update behavior \citep{peng2024eagle,peng2025rwkv7}. These models are closely related to our motivation because they show that efficient recurrent models benefit from stronger memory control. However, they still rely on compressed recurrent states, whereas our work studies how log-linear attention can adaptively weight multiple hierarchical memory levels.


Recent work has also revisited older recurrent memory ideas in modern sequence modeling. xLSTM extends the classical LSTM with exponential gating and modified memory structures, including scalar-memory and matrix-memory variants that can be scaled in residual language-model backbones \citep{beck2024xlstm}. Fast weight models provide an earlier view of adaptive memory, where a controller dynamically modifies rapidly changing weights during sequence processing \citep{schmidhuber1992fastweights}. These works reinforce that efficient sequence modeling depends not only on storing information, but also on how memory is written, updated, and accessed. Our method follows this motivation at the level of log-linear attention by making the memory-level weighting term more adaptive to the current input.


A growing body of work shows that recall is a central weakness of efficient sequence models. Zoology formalizes this issue through multi-query associative recall (MQAR), showing that much of the gap between attention and attention-free models comes from retrieving information mentioned earlier in context \citep{arora2023zoology}. H3 and BASED also study associative recall and the recall-throughput tradeoff, showing that smaller recurrent states often improve efficiency at the cost of precise retrieval \citep{fu2023h3,arora2024simple}. Other work highlights copying and long-context recall limitations in state-based models, while RULER evaluates long-context retrieval, tracing, and aggregation tasks \citep{jelassi2024illusion,hsieh2024ruler}. Compressive Transformer studies a related memory-compression problem, where older context is retained in compressed form but may lose task-relevant detail \citep{rae2019compressive}. Our evaluation follows this recall-focused direction by testing whether adaptive memory-level weighting improves associative recall in log-linear attention.


Recent hybrid architectures suggest another direction for efficient long-context modeling: combining recurrent memory with selective use of attention. Kimi Linear introduces a hybrid architecture built around Kimi Delta Attention, extending Gated DeltaNet with finer-grained gating while interleaving efficient linear attention layers with full-attention-style layers \citep{zhang2025kimilinear}. This supports the broader view that efficient long-context modeling depends not only on the presence of memory, but also on how memory is selected and controlled. Our work studies this issue at a finer level by modifying the $\lambda$ weights that determine how log-linear attention combines hierarchical memory levels.



Log-Linear Attention is the direct baseline for our work. Guo et al. address the fixed-state limitation of linear attention and state space models by replacing a single recurrent hidden state with a logarithmically growing hierarchy of hidden states \citep{guo2026loglinear}. This Fenwick-tree structure decomposes the past into power-of-two memory buckets, giving the model a middle ground between full attention, which stores all previous tokens, and linear attention, which compresses the entire prefix into one state. In this formulation, $\lambda$ controls how much each memory level contributes to the output. Rather than changing the Fenwick-tree memory structure or the asymptotic complexity, we focus on the underexplored parameterization of $\lambda$ and replace the original weakly input-dependent memory-level weighting with an MLP-parameterized version.



\section{Conclusion}
\label{sec:conclusion}

We identified a concrete limitation in log-linear attention: its memory-level
weighting term $\lambda$ is nearly input-independent in practice, applying
nearly uniform weights across hierarchy levels regardless of token content.
This rigidity causes high variance optimization and outright collapse under
high memory load, as demonstrated on multi-query associative recall with kv=32
where the baseline achieves only 2.9\% mean accuracy while our method holds
at 57.9\%. We addressed this with a lightweight two-layer MLP parameterization
using softplus activation, which produces per-token, per-level weights that
adapt to content rather than position. The modification preserves
$\mathcal{O}(T \log T)$ training complexity and $\mathcal{O}(\log T)$ decoding
memory exactly, adding fewer than $0.007\%$ additional parameters. Across
multi-query associative recall, selective copying at seq=1024, and WikiText-103
language modeling, adaptive $\lambda$ consistently outperforms the baseline,
with the largest gains precisely where baseline $\lambda$ degrades or collapses.
Lambda visualizations confirm that the MLP learns structured, content-dependent
routing through the Fenwick-tree hierarchy rather than the near-uniform
weighting of the baseline. We hope this work encourages further exploration of
adaptive memory control in log-linear and hierarchical attention architectures.

\bibliographystyle{plainnat}
\bibliography{references}

\appendix
\setlength{\floatsep}{2pt plus 1pt minus 1pt}
\setlength{\textfloatsep}{4pt plus 1pt minus 1pt}
\setlength{\intextsep}{2pt plus 1pt minus 1pt}
\vspace{-0.5em}
\section{Additional Experimental Details}
\label{appendix:details}
\vspace{-0.5em}
\subsection{Model Configuration}

All models use a two-layer HAttention architecture with hidden dimension 64,
2 attention heads, and head dimension 32. For MQAR and selective copying
experiments, the vocabulary size is set to match the task (128 for MQAR,
64 for selective copying). For language modeling, we use hidden size 512,
6 transformer layers, 4 attention heads, head dimension 64, and GPT-2
vocabulary size 50,257. MLP-$\lambda$ variants use a two-layer MLP with
hidden dimension $d_h = 64$ and GELU activation. $\mathbf{W}_1$ is
initialized with Xavier uniform initialization, $\mathbf{W}_2$ is
initialized to zero, and the scalar bias is set to $0.54$ so that
$\mathrm{softplus}(0.54) \approx 1.0$ at initialization.
\vspace{-0.5em}
\subsection{Training Setup}

\paragraph{MQAR.} Models are trained for 5000 steps with the Adam optimizer
at learning rate $10^{-3}$ and batch size 64. No learning rate schedule is
used. We vary the number of key-value pairs $k \in \{4, 8, 16, 32\}$ and
sequence lengths $\in \{256, 512\}$, running 3--5 random seeds per
configuration.

\paragraph{Selective copying.} Models are trained for 30,000 steps with the
Adam optimizer at learning rate $10^{-3}$ and batch size 64. Sequence lengths
are $\in \{256, 512, 1024\}$ with 16 target tokens. Five seeds are used for
seq=256 and seq=512; three seeds for seq=1024.

\paragraph{Language modeling.} Models are trained for 40,000 steps on
WikiText-103 using the AdamW optimizer with learning rate $3 \times 10^{-4}$,
weight decay 0.1, batch size 8, and sequence length 512. A linear warmup
over 500 steps is followed by cosine decay. The GPT-2 tokenizer is used
with vocabulary size 50,257. Validation perplexity is reported on the
WikiText-103 validation split.

\vspace{-0.5em}
\subsection{Hardware and Implementation}

All experiments were run on a single NVIDIA A100 SXM4 80GB GPU. Models are
implemented in PyTorch. The HAttention backbone follows the implementation
of \citet{guo2026loglinear}. Training uses bfloat16 mixed precision and
TF32 for matrix multiplications. MQAR and selective copying experiments
each take approximately 5--15 minutes per seed. Language modeling runs
take approximately 3--4 hours per seed at 40,000 steps.

\subsection{Compute Budget}

The total compute budget for all reported experiments is approximately
100--120 GPU-hours on a single A100 80GB. This includes MQAR experiments across all kv and sequence length settings (roughly 40 GPU-hours), selective copying across all sequence lengths and seeds (roughly 30 GPU-hours), and language modeling runs (roughly 12 GPU-hours). Additional preliminary experiments and ablations not reported in the paper account for the
remainder.

\subsection{MLP Hidden Dimension Ablation}
\label{appendix:dim_ablation}

Table~\ref{tab:dim_ablation} reports MQAR accuracy across 11 random seeds for 
three MLP hidden dimension choices: $d_h \in \{32, 64, 128\}$. All three 
dimensions achieve comparable convergence rates when they converge, with 
per-seed averages of approximately 99.2\%, 99.2\%, and 99.1\% respectively 
among converged seeds. The occasional non-convergent seeds (3.3\%--3.4\%) 
reflect the general seed sensitivity of the MQAR task rather than 
dimension-specific instability, as they appear across all three settings. 
Based on this ablation, we select $d_h = 64$ as the minimal sufficient 
hidden dimension --- it matches the expressivity of larger dimensions while 
keeping parameter overhead minimal.

\begin{table}[H]
\caption{MQAR accuracy (\%) across 11 seeds for MLP hidden dimensions 
$d_h \in \{32, 64, 128\}$. Converged seeds achieve $\sim$99\% regardless 
of dimension, confirming $d_h=64$ is sufficient.}
\label{tab:dim_ablation}
\centering
\small
\begin{tabular}{cccc}
\toprule
\textbf{Seed} & \textbf{Dim 32} & \textbf{Dim 64} & \textbf{Dim 128} \\
\midrule
0  & 99.1\% & 3.3\%  & 3.3\%  \\
1  & 3.3\%  & 3.2\%  & 99.2\% \\
2  & 99.0\% & 99.1\% & 99.0\% \\
3  & 99.5\% & 99.3\% & 99.2\% \\
4  & 99.3\% & 99.2\% & 3.3\%  \\
5  & 99.1\% & 99.2\% & 99.2\% \\
6  & 99.2\% & 99.7\% & 99.1\% \\
7  & 99.3\% & 99.3\% & 99.3\% \\
8  & 3.4\%  & 99.0\% & 98.2\% \\
9  & 99.2\% & 99.3\% & 99.5\% \\
10 & 99.2\% & 99.3\% & 96.4\% \\
\midrule
\textbf{Avg (converged)} & \textbf{$\sim$99.2\%} & \textbf{$\sim$99.2\%} & \textbf{$\sim$99.1\%} \\
\bottomrule
\end{tabular}
\end{table}

\vspace{-0.5em}
\section{Additional Results}
\label{appendix:results}

\subsection{Per-Seed Results}

Table~\ref{tab:per_seed_selective} reports per-configuration mean and
standard deviation for the selective copying task across all sequence
lengths and modes. Results at seq=1024 use 3 seeds; all others use 5 seeds.

\begin{table}[H]
  \caption{Per-configuration selective copying accuracy (\%).
  Mean $\pm$ std over seeds.}
  \label{tab:per_seed_selective}
  \centering
  \begin{tabular}{llcc}
    \toprule
    Mode & Seq len & Seeds & Mean $\pm$ Std \\
    \midrule
    Baseline $\lambda$       & 256  & 5 & $52.6 \pm 3.5$ \\
    Baseline $\lambda$       & 512  & 5 & $54.6 \pm 3.4$ \\
    Baseline $\lambda$       & 1024 & 3 & $40.6 \pm 19.8$ \\
    \midrule
    MLP-$\lambda$ (softmax)  & 256  & 5 & $54.3 \pm 5.8$ \\
    MLP-$\lambda$ (softmax)  & 512  & 5 & $50.3 \pm 0.8$ \\
    MLP-$\lambda$ (softmax)  & 1024 & 3 & $52.6 \pm 2.5$ \\
    \midrule
    MLP-$\lambda$ (softplus) & 256  & 5 & $53.1 \pm 2.2$ \\
    MLP-$\lambda$ (softplus) & 512  & 5 & $51.9 \pm 0.8$ \\
    MLP-$\lambda$ (softplus) & 1024 & 3 & $\mathbf{54.8 \pm 2.3}$ \\
    \bottomrule
  \end{tabular}
\end{table}

The most notable pattern is the dramatically higher variance of the fixed
baseline at seq=1024 ($\pm$19.8) compared to both MLP variants
($\pm$2.3--2.5), indicating seed-dependent collapse rather than consistent
learning. This mirrors the instability observed in MQAR at high kv counts.

\vspace{-0.5em}
\subsection{Training Dynamics}
Learning curves for selective copying are shown in
Figure~\ref{fig:selective} in the main paper. At seq=256 and seq=512,
all three methods converge at comparable rates and final accuracies,
with MLP-softmax showing slightly higher variance across seeds at seq=256.
At seq=1024, the baseline plateaus near 40\% well before 30,000
steps, while both MLP variants continue to improve steadily throughout
training, converging near 52--55\%. This divergence in training dynamics
is consistent with the hypothesis that adaptive $\lambda$ provides more
stable gradient signal under high memory pressure, preventing the
optimization collapse seen in the fixed parameterization.

\subsection{Lambda Visualization}
\label{appendix:lambda_visualization}

\subsubsection{Fixed-\texorpdfstring{$\lambda$}{lambda} Heatmaps Across MQAR Settings}
\label{appendix:fixed_lambda_mqar}

We first visualize the learned baseline-$\lambda$ weights on MQAR at sequence
length 256 for two association densities, kv=16 and kv=32. Since the fixed
baseline assigns one learned weight to each hierarchy level, these heatmaps do
not vary over token position. Each plot shows the global memory-level profile
learned by a layer, with columns corresponding to Fenwick-tree hierarchy levels
and rows corresponding to attention heads.

\begin{figure}[htbp]
  \centering
  \begin{minipage}{0.49\textwidth}
    \centering
    \includegraphics[width=0.8\linewidth]{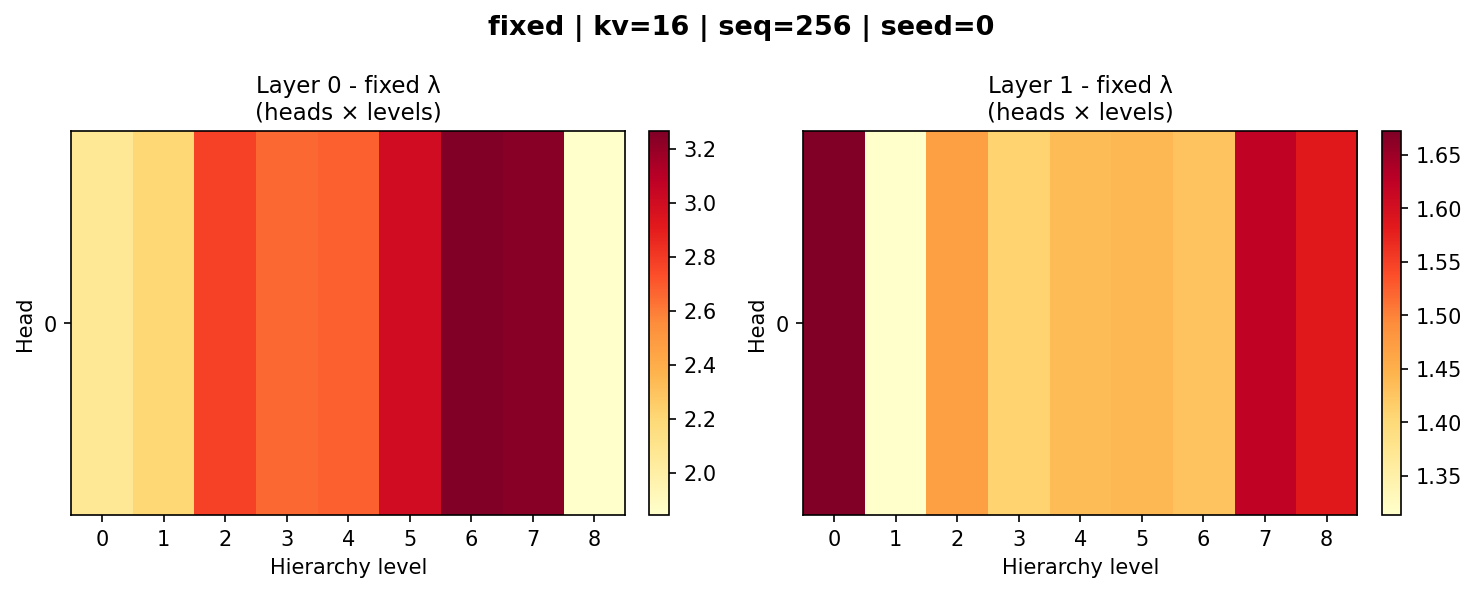}
  \end{minipage}
  \hfill
  \begin{minipage}{0.49\textwidth}
    \centering
    \includegraphics[width=0.8\linewidth]{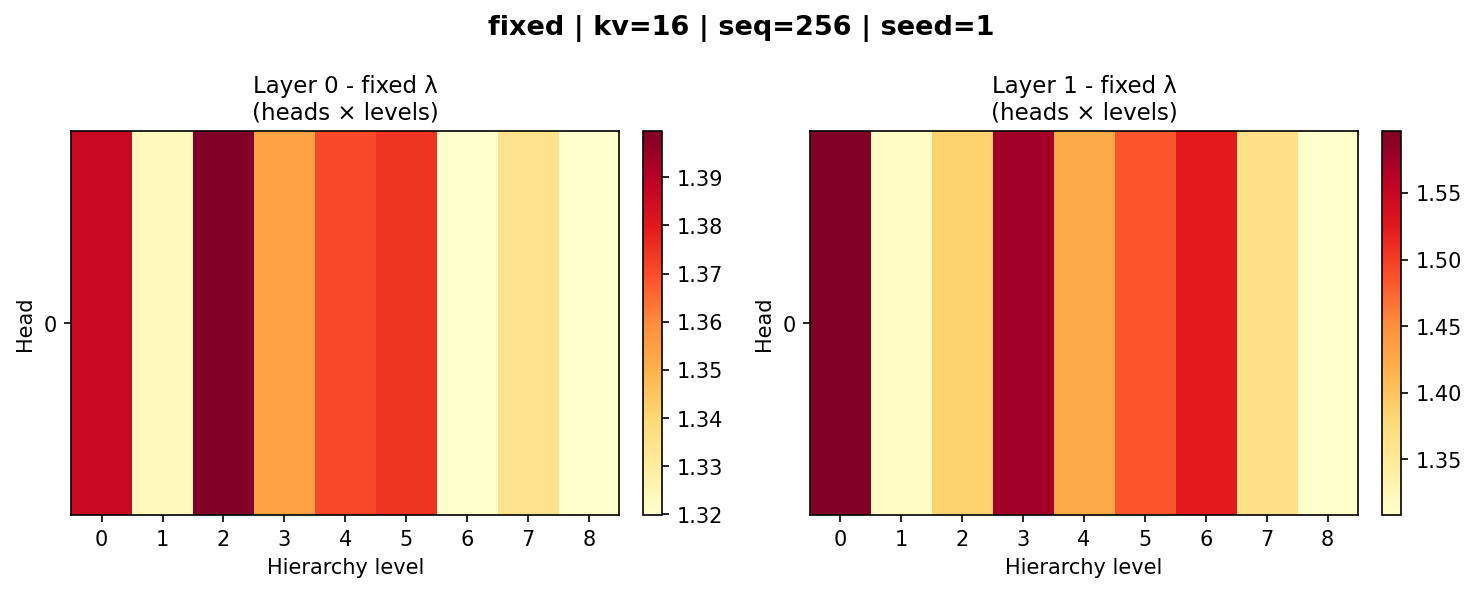}
  \end{minipage}
  \vspace{-0.5em}
  \begin{minipage}{0.49\textwidth}
    \centering
    \includegraphics[width=0.8\linewidth]{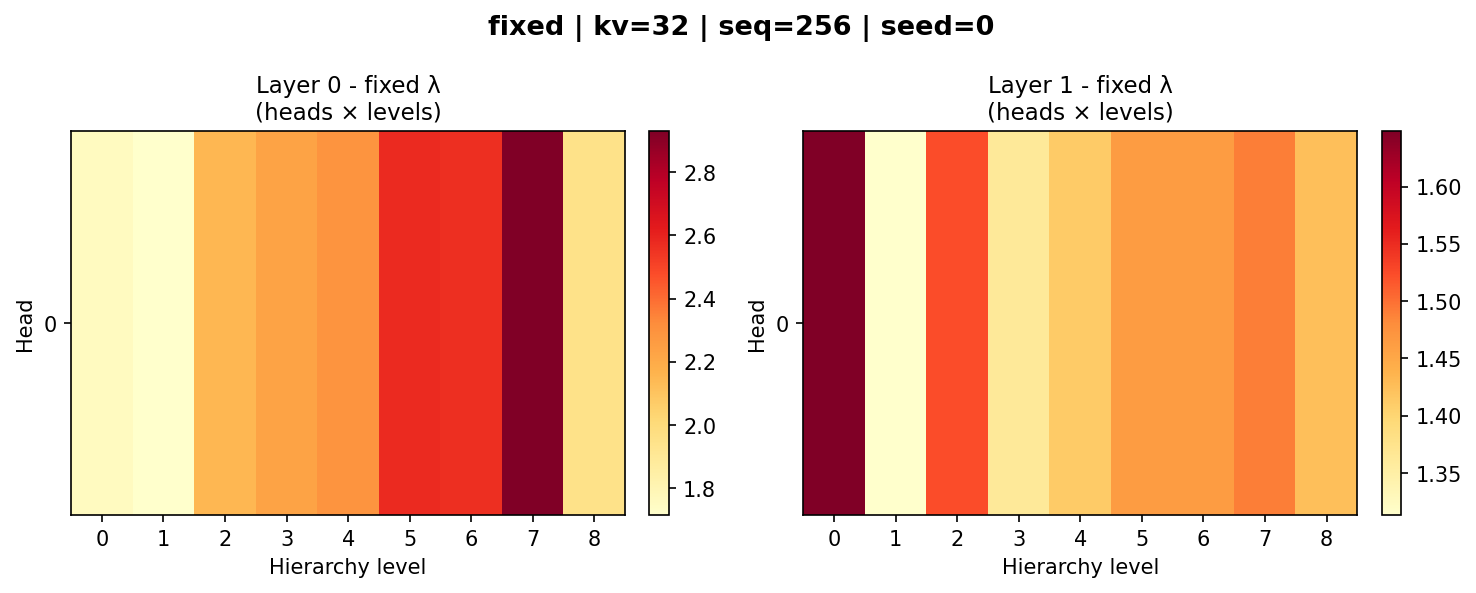}
  \end{minipage}
  \hfill
  \begin{minipage}{0.49\textwidth}
    \centering
    \includegraphics[width=0.8\linewidth]{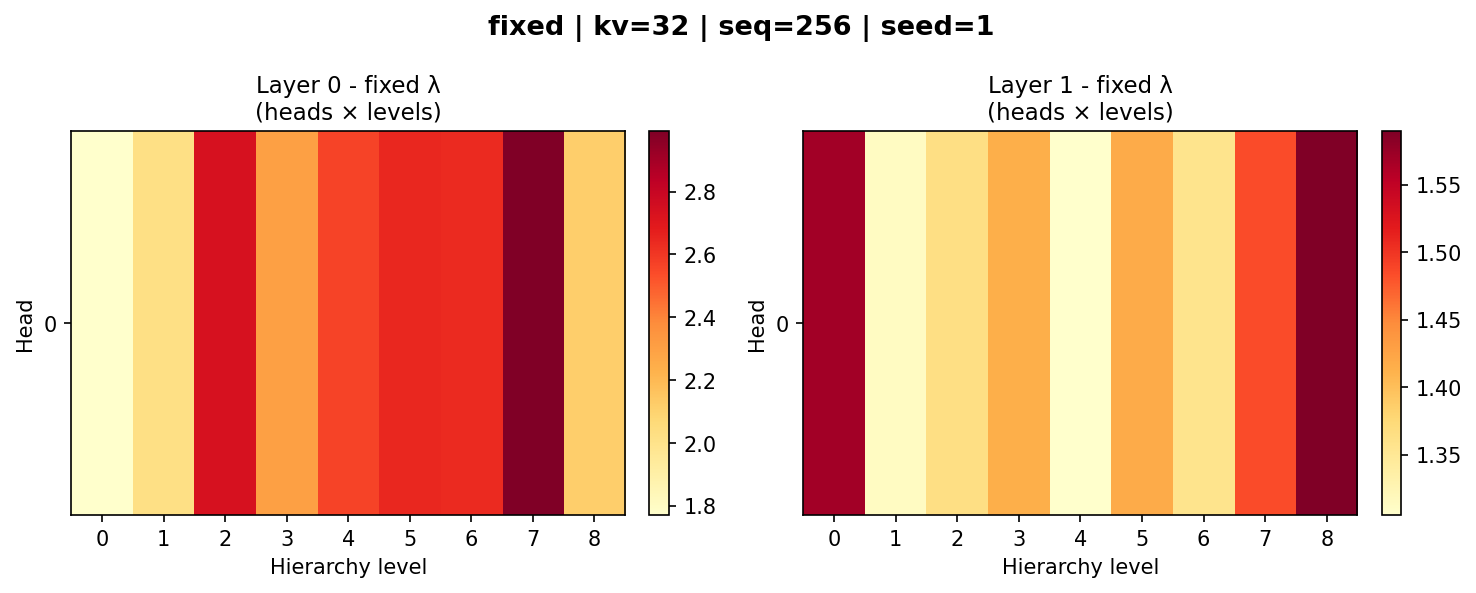}
  \end{minipage}
  \vspace{-0.5em}
  \caption{Baseline-$\lambda$ heatmaps on MQAR for kv=16 and kv=32 at sequence
  length 256 across two random seeds. The baseline learns non-uniform
  memory-level preferences, but these are static and do not change across token
  positions.}
  \label{fig:appendix_fixed_lambda_mqar}
\end{figure}

Figure~\ref{fig:appendix_fixed_lambda_mqar} shows that the baseline learns
non-uniform preferences over the hierarchy, differing between Layer 0 and
Layer 1, with larger weight often placed on a small subset of middle or deeper
levels. These patterns are broadly consistent across kv=16 and kv=32, though
the exact scale and preferred levels vary somewhat across seeds. Fixed
$\lambda$ can learn a meaningful global memory strategy per layer, but applies
it identically at every token --- the key limitation that MLP-parameterized
variants address.

\subsubsection{Token-Level MLP-Softplus \texorpdfstring{$\lambda$}{lambda} Heatmaps on MQAR}
\label{appendix:mlp_softplus_token_mqar}

We next visualize the per-token $\lambda$ weights produced by the MLP-softplus
parameterization on MQAR. Unlike the baseline, this variant computes
memory-level weights as a function of the current token representation, so the
heatmaps vary along the token-position axis. Each row corresponds to a
Fenwick-tree hierarchy level, and each column to a token position.

\begin{figure}[htbp]
  \centering
  \begin{minipage}{0.49\textwidth}
    \centering
    \includegraphics[width=\linewidth]{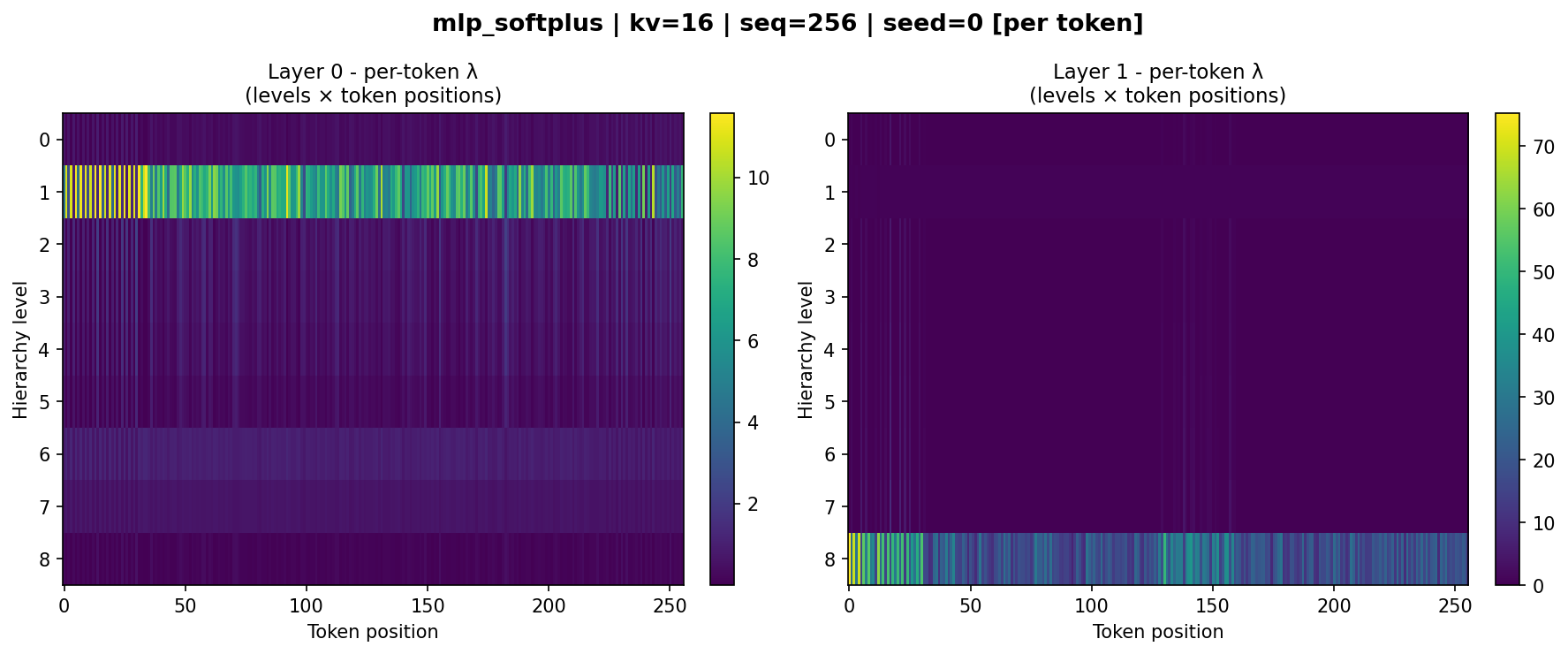}
  \end{minipage}
  \hfill
  \begin{minipage}{0.49\textwidth}
    \centering
    \includegraphics[width=\linewidth]{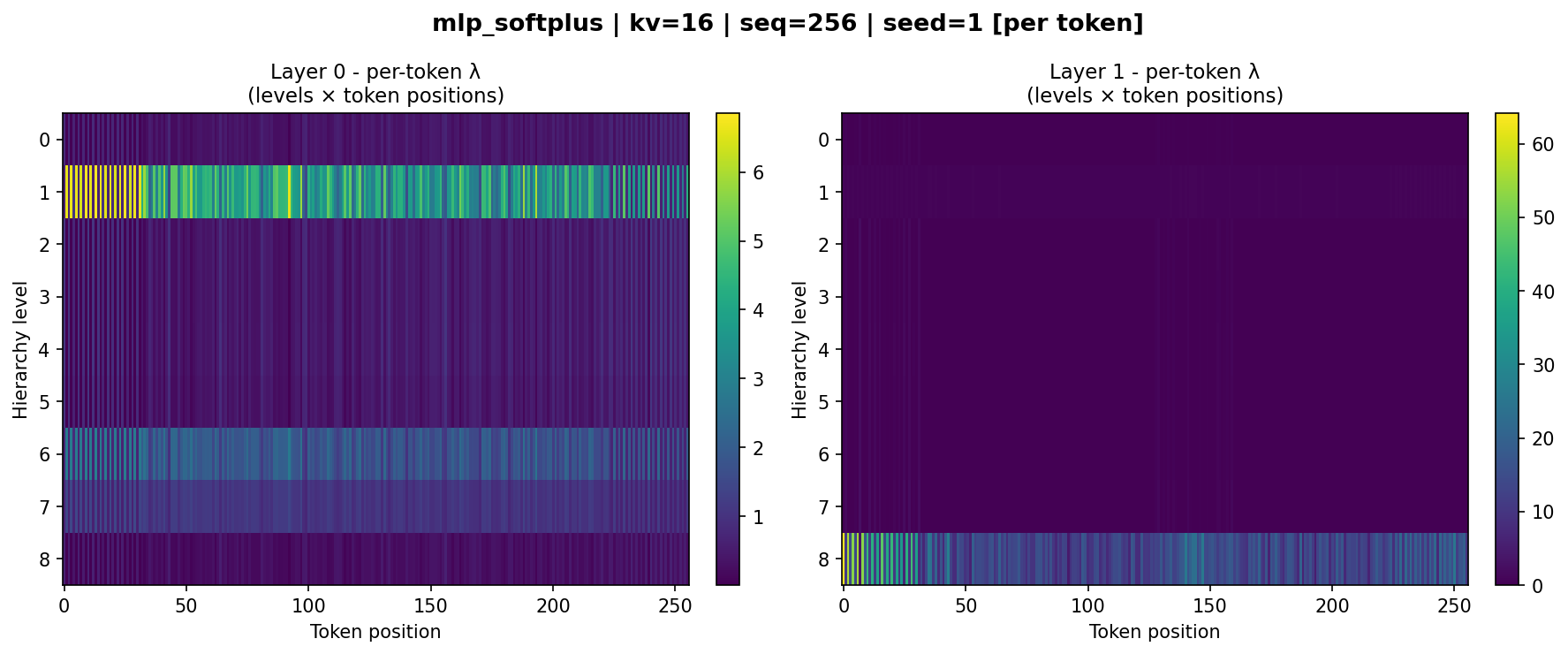}
  \end{minipage}
  \vspace{-0.5em}
  \begin{minipage}{0.49\textwidth}
    \centering
    \includegraphics[width=\linewidth]{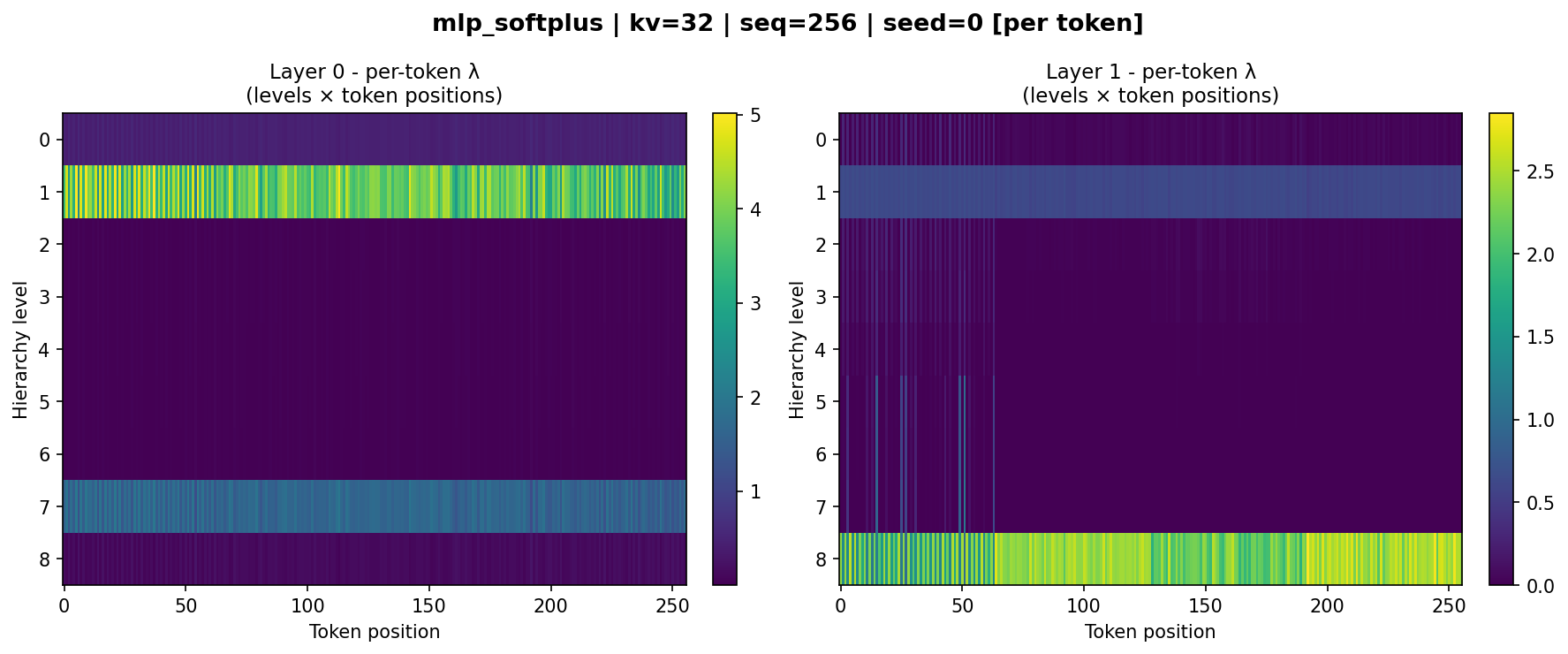}
  \end{minipage}
  \hfill
  \begin{minipage}{0.49\textwidth}
    \centering
    \includegraphics[width=\linewidth]{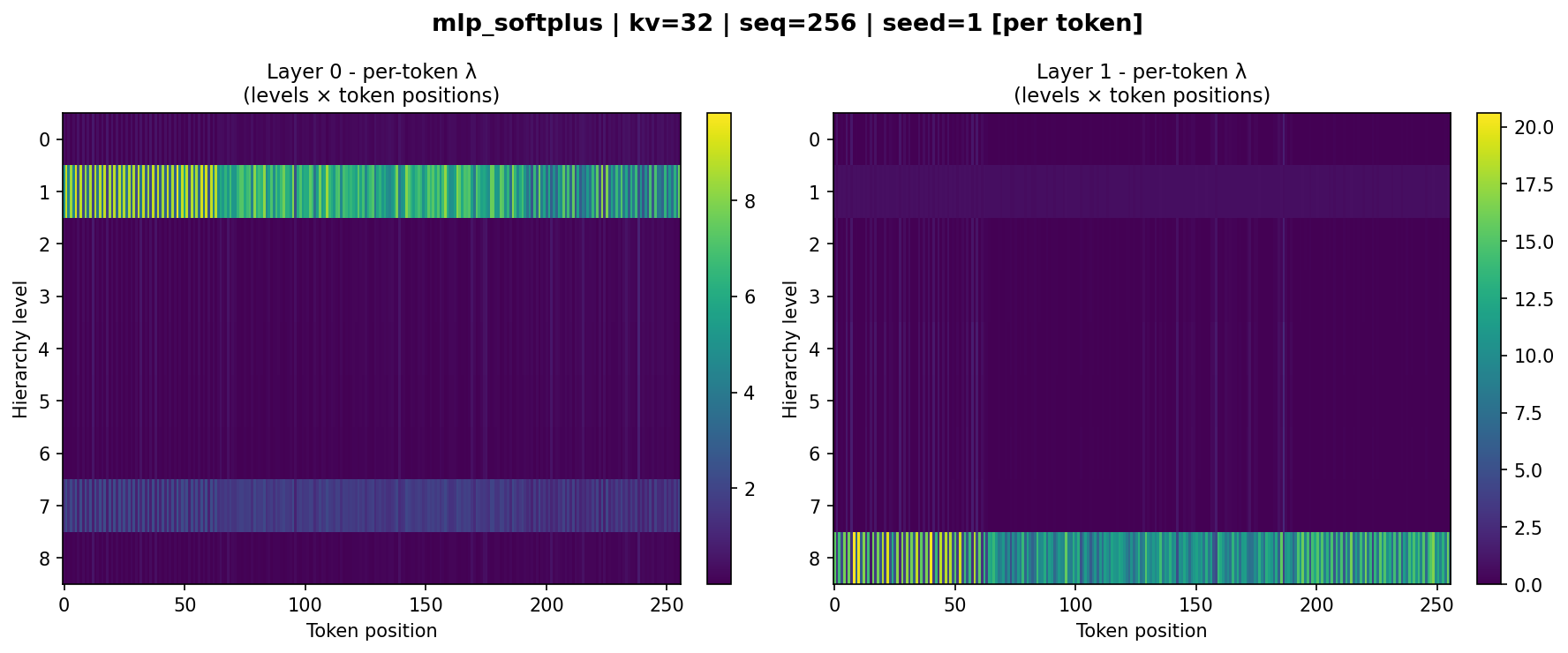}
  \end{minipage}
  \vspace{-0.5em}
  \caption{Token-level MLP-softplus $\lambda$ heatmaps on MQAR for kv=16 and
  kv=32 at sequence length 256 across two random seeds. Unlike the baseline,
  the strength of level preferences varies across token positions.}
  \label{fig:appendix_mlp_softplus_token_mqar}
\end{figure}

Figure~\ref{fig:appendix_mlp_softplus_token_mqar} shows that MLP-softplus
learns structured token-dependent weighting patterns. For both kv=16 and
kv=32, the model concentrates weight on a small subset of hierarchy levels, but
the strength of these levels changes across token positions. Layer 0 tends to
emphasize lower or intermediate levels, while Layer 1 often places more weight
on deeper levels, suggesting the two layers use the memory hierarchy
differently.

\subsubsection{Average MLP-Softplus \texorpdfstring{$\lambda$}{lambda} Heatmaps on MQAR}
\label{appendix:mlp_softplus_avg_mqar}

We also average the MLP-softplus $\lambda$ weights over tokens and batch
examples to summarize global memory-level preferences. This view removes
token-wise variation but makes layer-wise structure easier to see.

\begin{figure}[htbp]
  \centering
  \begin{minipage}{0.49\textwidth}
    \centering
    \includegraphics[width=\linewidth]{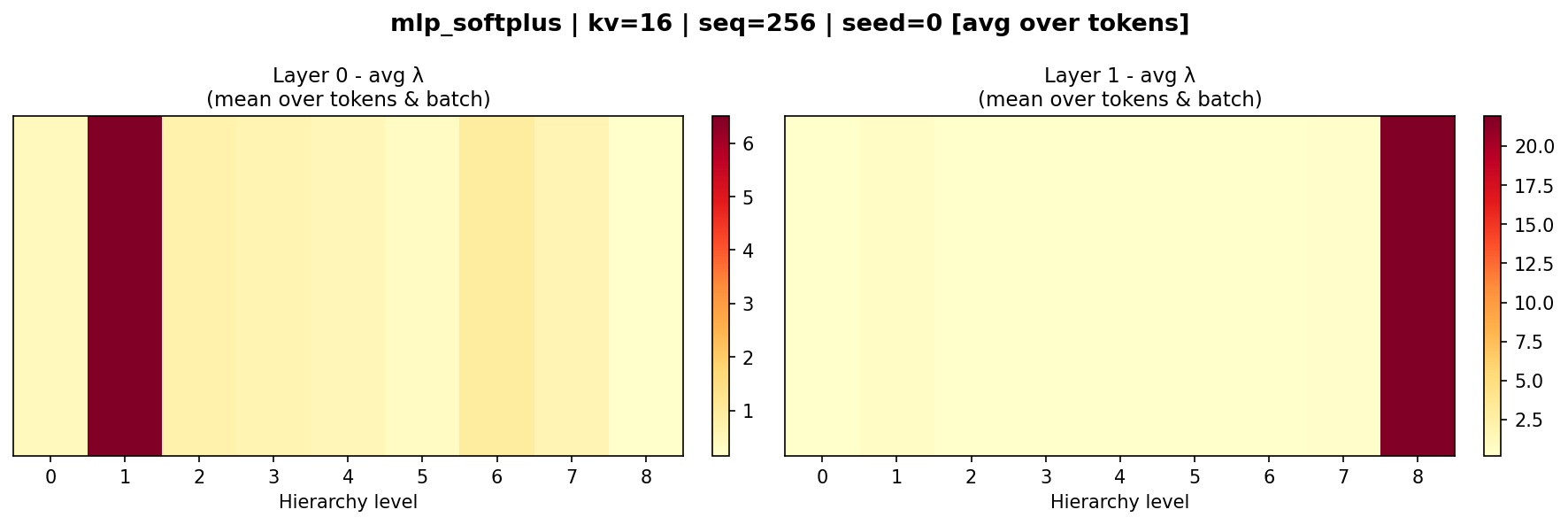}
  \end{minipage}
  \hfill
  \begin{minipage}{0.49\textwidth}
    \centering
    \includegraphics[width=\linewidth]{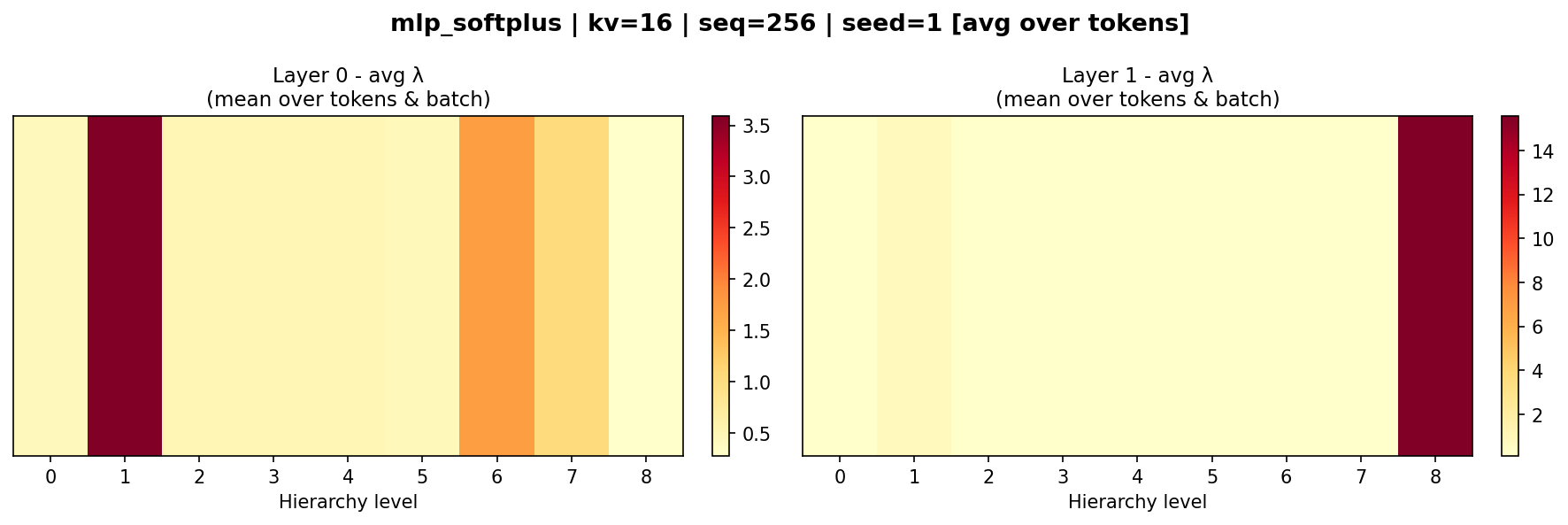}
  \end{minipage}
  \vspace{-0.5em}
  \begin{minipage}{0.49\textwidth}
    \centering
    \includegraphics[width=\linewidth]{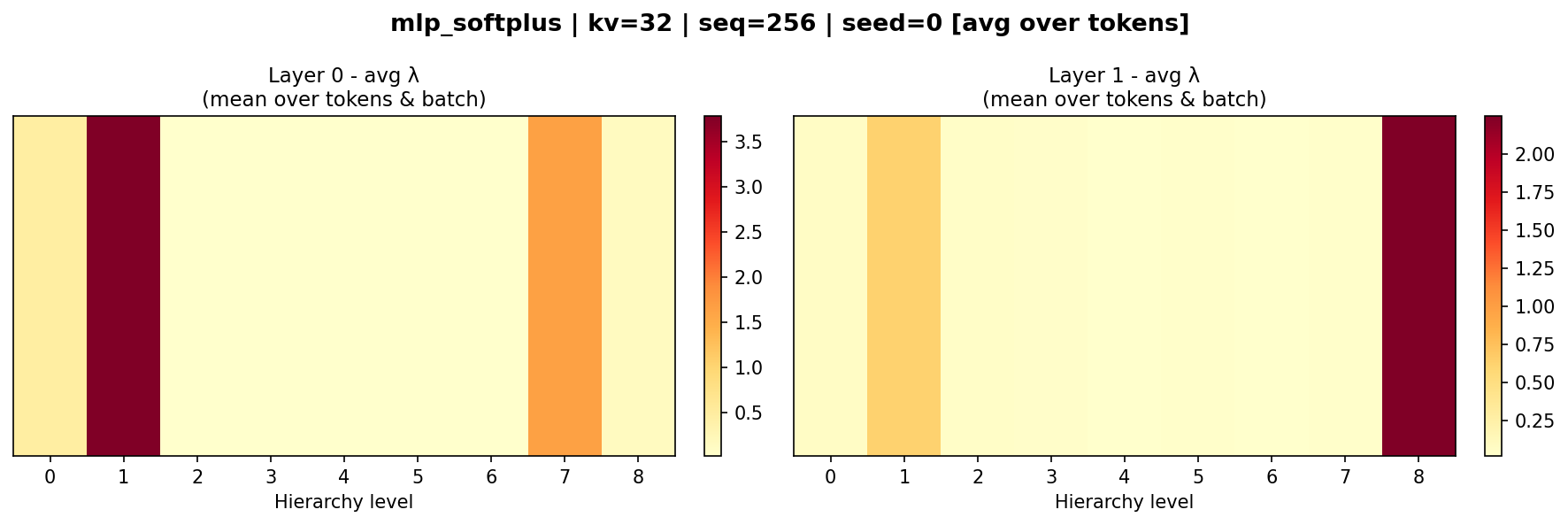}
  \end{minipage}
  \hfill
  \begin{minipage}{0.49\textwidth}
    \centering
    \includegraphics[width=\linewidth]{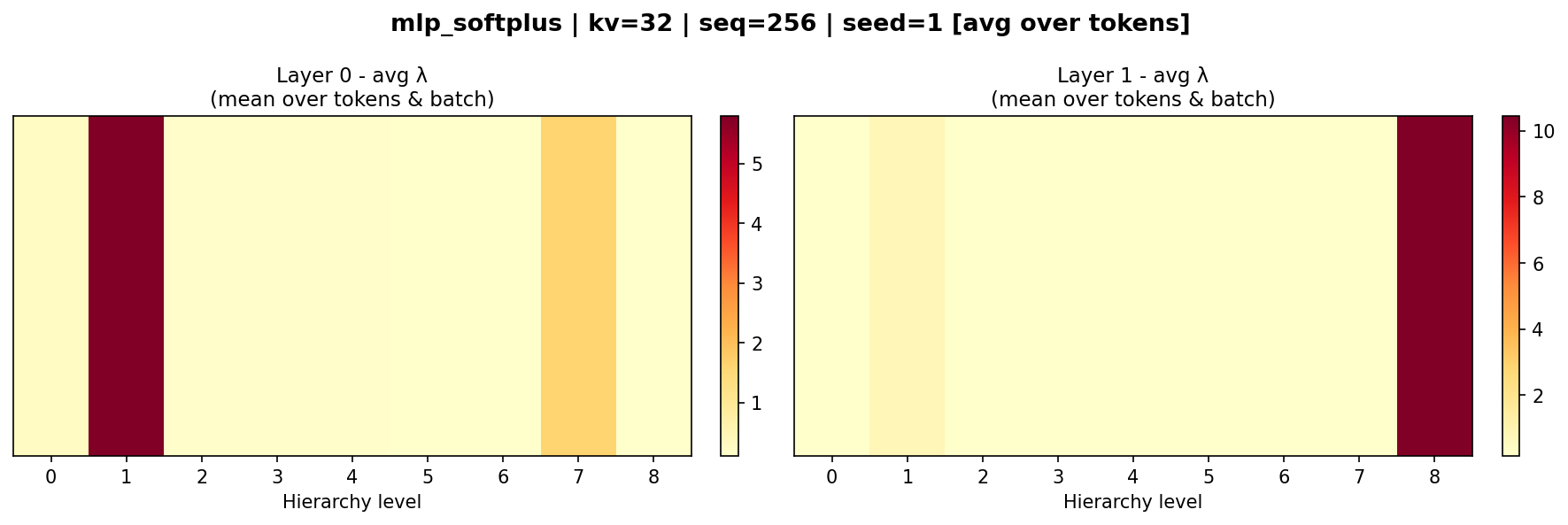}
  \end{minipage}
  \vspace{-0.5em}
  \caption{Average MLP-softplus $\lambda$ heatmaps on MQAR for kv=16 and kv=32
  at sequence length 256 across two random seeds. Averaging highlights the
  sparse, layer-specific memory-level preferences learned by MLP-softplus.}
  \label{fig:appendix_mlp_softplus_avg_mqar}
\end{figure}

Figure~\ref{fig:appendix_mlp_softplus_avg_mqar} shows that MLP-softplus learns
sparse, layer-specific profiles over the Fenwick hierarchy. Layer 0 places most
average weight on lower or mid-to-deep levels, while Layer 1 consistently
weights the deepest level most heavily. The separation between layers is stable
across seeds. Together with the token-level heatmaps, this shows MLP-softplus
learns both a global preference over memory scales and token-dependent
deviations from that preference.

\subsubsection{Token-Level MLP-Softmax \texorpdfstring{$\lambda$}{lambda} Heatmaps on MQAR}
\label{appendix:mlp_softmax_token_mqar}

For completeness, we visualize token-level $\lambda$ weights for the
MLP-softmax variant. Because softmax normalizes across hierarchy levels, it
encourages stronger competition between levels and often produces sharper level
preferences than MLP-softplus.

\begin{figure}[htbp]
  \centering
  \begin{minipage}{0.49\textwidth}
    \centering
    \includegraphics[width=\linewidth]{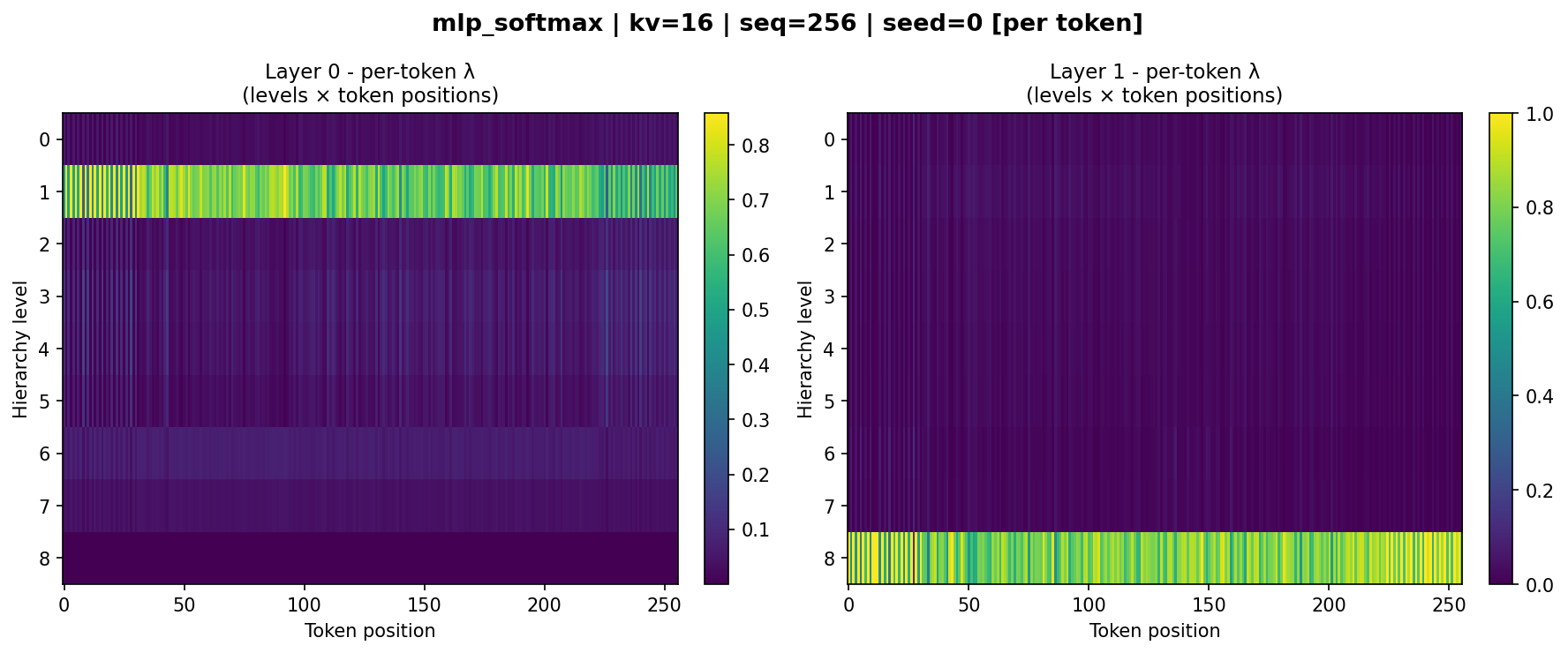}
  \end{minipage}
  \hfill
  \begin{minipage}{0.49\textwidth}
    \centering
    \includegraphics[width=\linewidth]{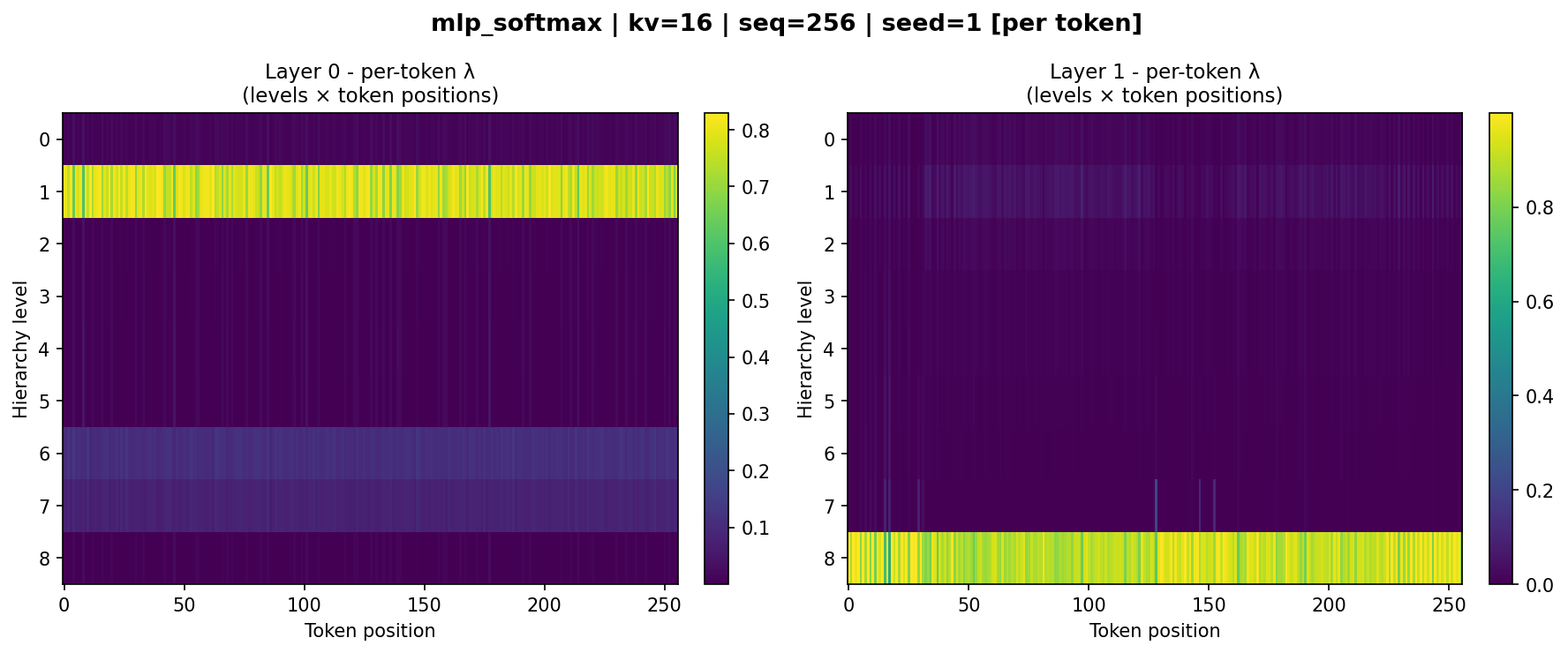}
  \end{minipage}
  \vspace{-0.5em}
  \begin{minipage}{0.49\textwidth}
    \centering
    \includegraphics[width=\linewidth]{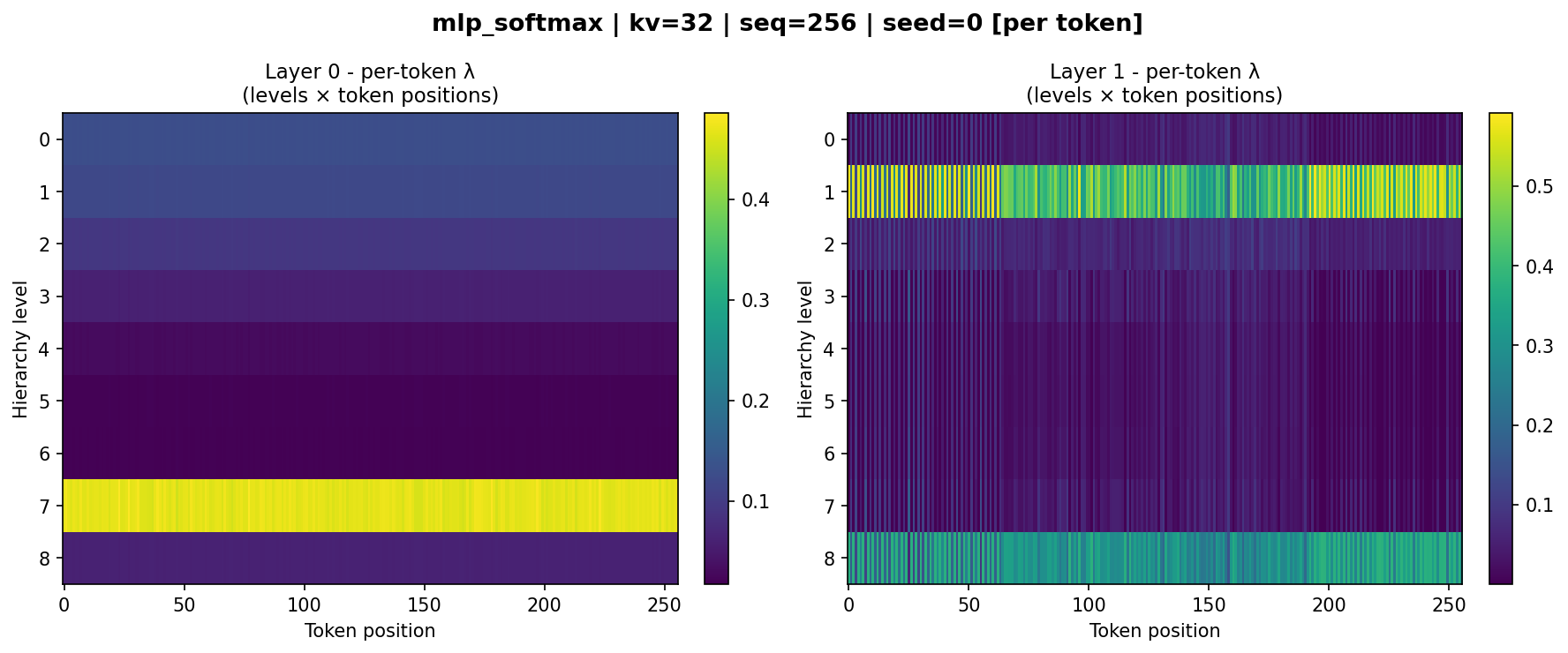}
  \end{minipage}
  \hfill
  \begin{minipage}{0.49\textwidth}
    \centering
    \includegraphics[width=\linewidth]{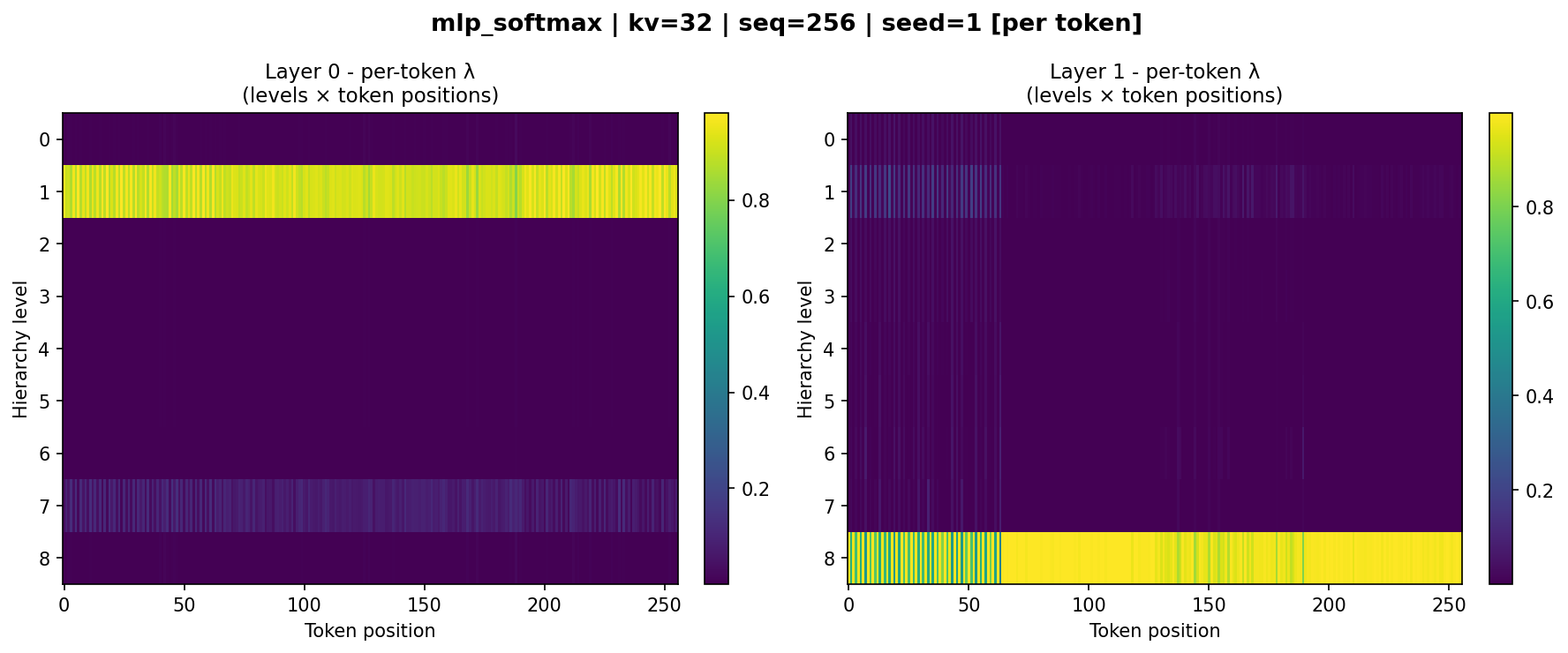}
  \end{minipage}
  \vspace{-0.5em}
  \caption{Token-level MLP-softmax $\lambda$ heatmaps on MQAR for kv=16 and
  kv=32 at sequence length 256 across two random seeds. Softmax normalization
  often leads to sharper and more seed-sensitive level preferences than
  MLP-softplus.}
  \label{fig:appendix_mlp_softmax_token_mqar}
\end{figure}

Figure~\ref{fig:appendix_mlp_softmax_token_mqar} shows that MLP-softmax also
learns structured memory-level weighting. At kv=16, one layer places most
weight on a lower level while the other weights the deepest level. At kv=32,
the active levels vary more noticeably across seeds, consistent with softmax
normalization shifting mass between competing levels.

\subsubsection{Selective Copying Lambda Heatmaps}
\label{appendix:lambda_selective_copy}

We also visualize $\lambda$ weights on selective copying to test whether the
fixed-versus-adaptive behavior observed on MQAR appears in a different
synthetic memory task. In selective copying, the model must preserve earlier
tokens and reproduce them near the end of the sequence. The red dashed line in
the token-level heatmaps marks the beginning of the copy target region. We show
tok=16 results across sequence lengths 256, 512, and 1024.

\begin{figure}[htbp]
\centering
\begin{minipage}{0.49\textwidth}
    \centering
    \includegraphics[width=\linewidth]{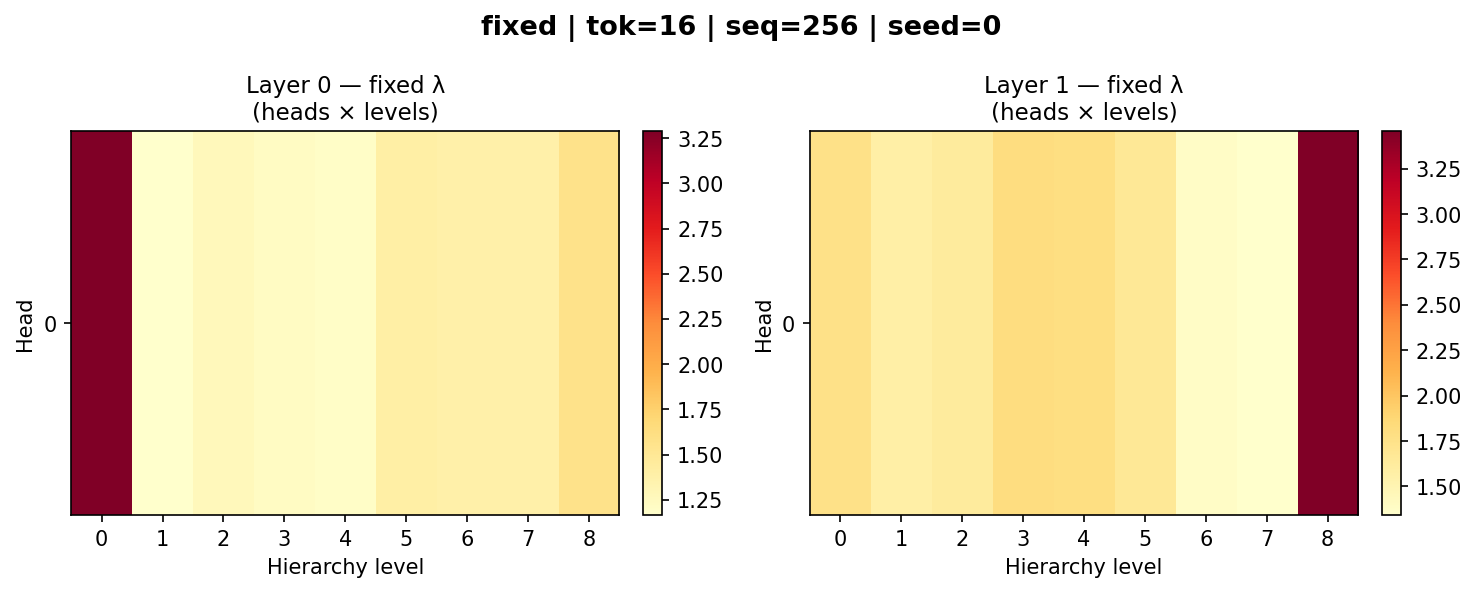}
\end{minipage}
\hfill
\begin{minipage}{0.49\textwidth}
    \centering
    \includegraphics[width=\linewidth]{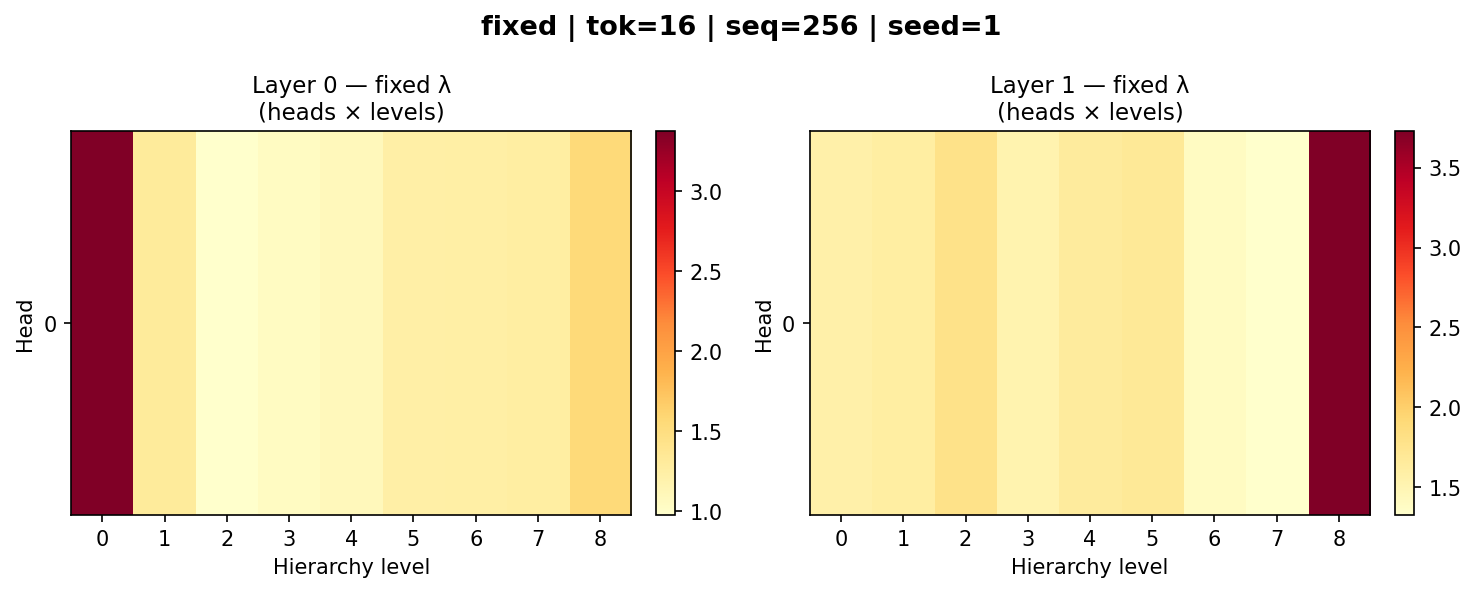}
\end{minipage}
\vspace{0.2em}
\begin{minipage}{0.49\textwidth}
    \centering
    \includegraphics[width=\linewidth]{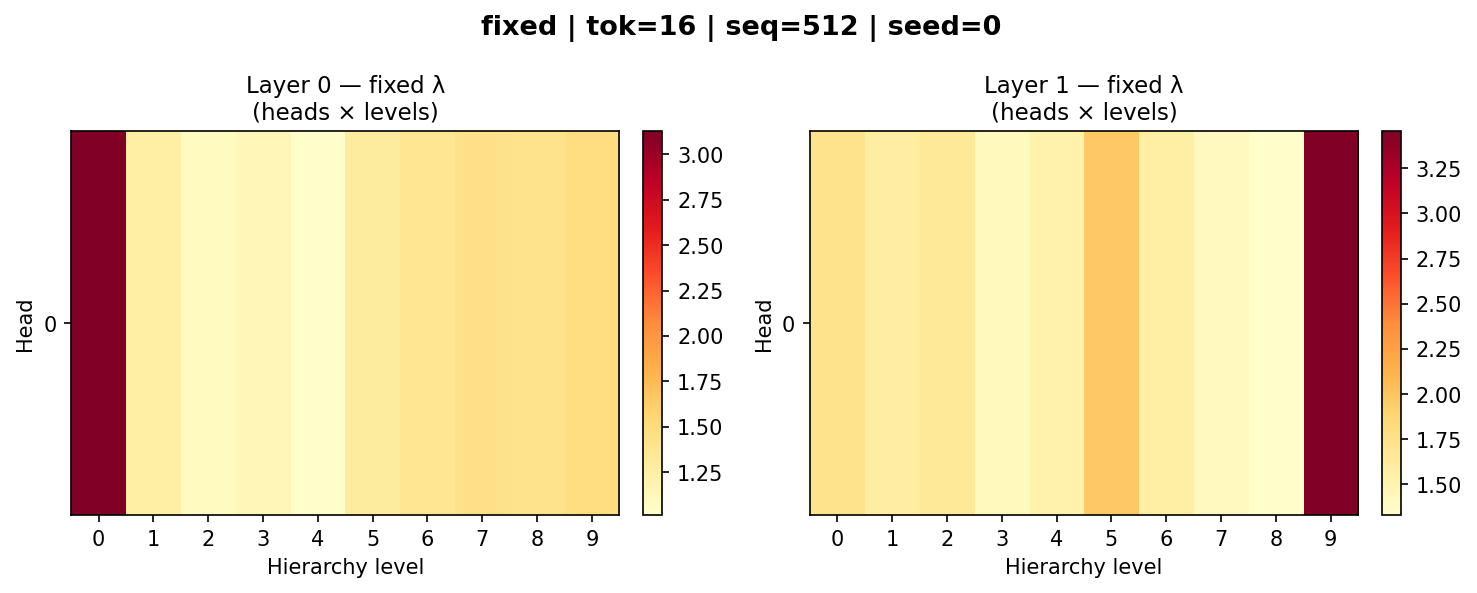}
\end{minipage}
\hfill
\begin{minipage}{0.49\textwidth}
    \centering
    \includegraphics[width=\linewidth]{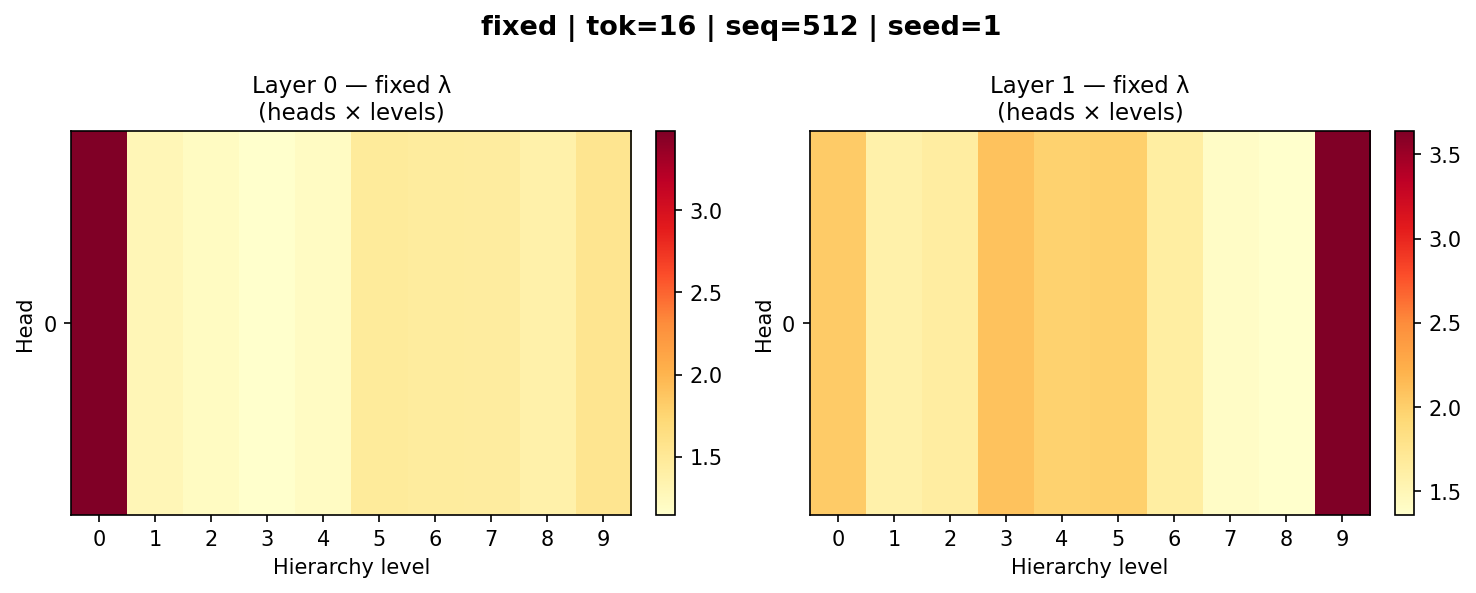}
\end{minipage}
\vspace{0.2em}
\begin{minipage}{0.49\textwidth}
    \centering
    \includegraphics[width=\linewidth]{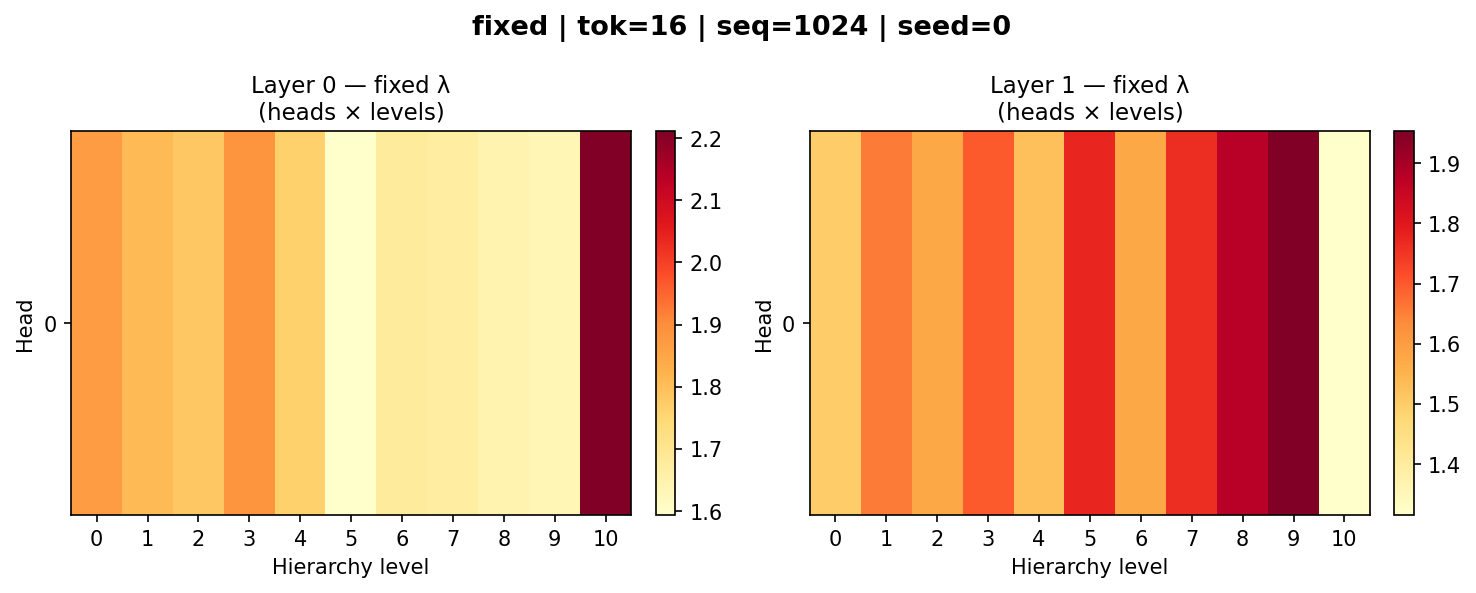}
\end{minipage}
\hfill
\begin{minipage}{0.49\textwidth}
    \centering
    \includegraphics[width=\linewidth]{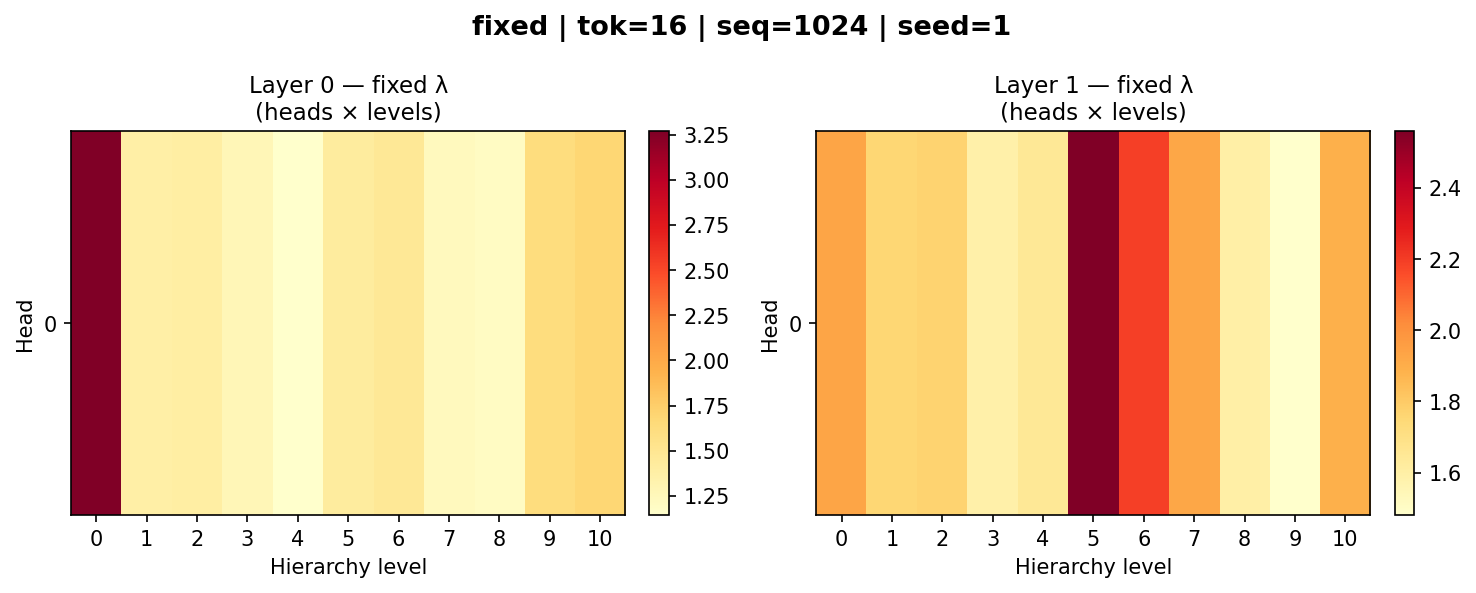}
\end{minipage}
\caption{Baseline-$\lambda$ heatmaps on selective copying for tok=16 across
sequence lengths 256, 512, and 1024. Each row shows two random seeds for one
sequence length. The baseline learns non-uniform layer-wise preferences
over hierarchy levels, but these preferences are static and do not vary across
token positions.}
\label{fig:sc_fixed_all}
\end{figure}

Figure~\ref{fig:sc_fixed_all} shows the Baseline-$\lambda$ baseline across all
three sequence lengths. The baseline learns layer-specific preferences
over Fenwick-tree hierarchy levels, but these preferences remain static with
respect to token position. This behavior is consistent with the MQAR fixed
heatmaps: the model can learn a global memory profile for each layer, but it
cannot change that profile as the sequence approaches the copy target region.
The longer sequence lengths introduce additional hierarchy levels, but do not
change the static nature of the fixed parameterization.

\begin{figure}[htbp]
\centering
\begin{minipage}{0.49\textwidth}
    \centering
    \includegraphics[width=\linewidth]{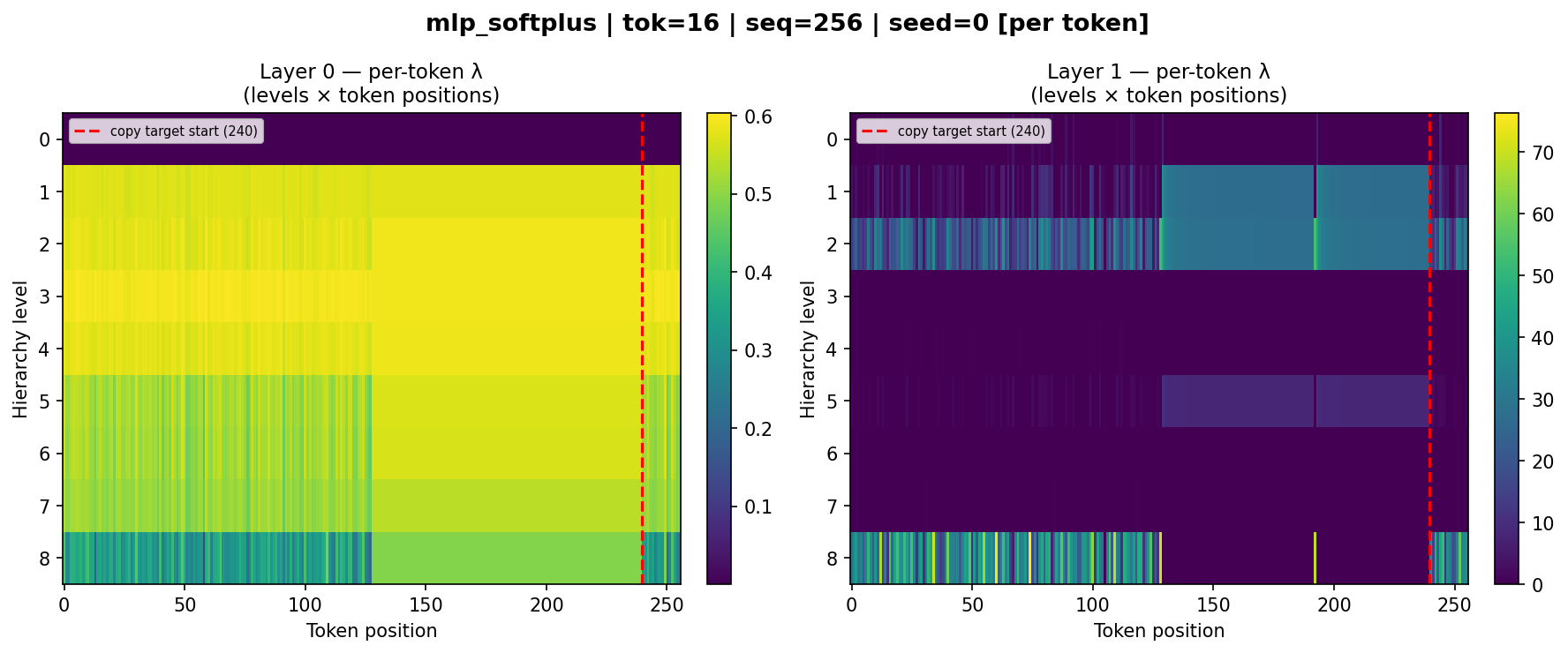}
\end{minipage}
\hfill
\begin{minipage}{0.49\textwidth}
    \centering
    \includegraphics[width=\linewidth]{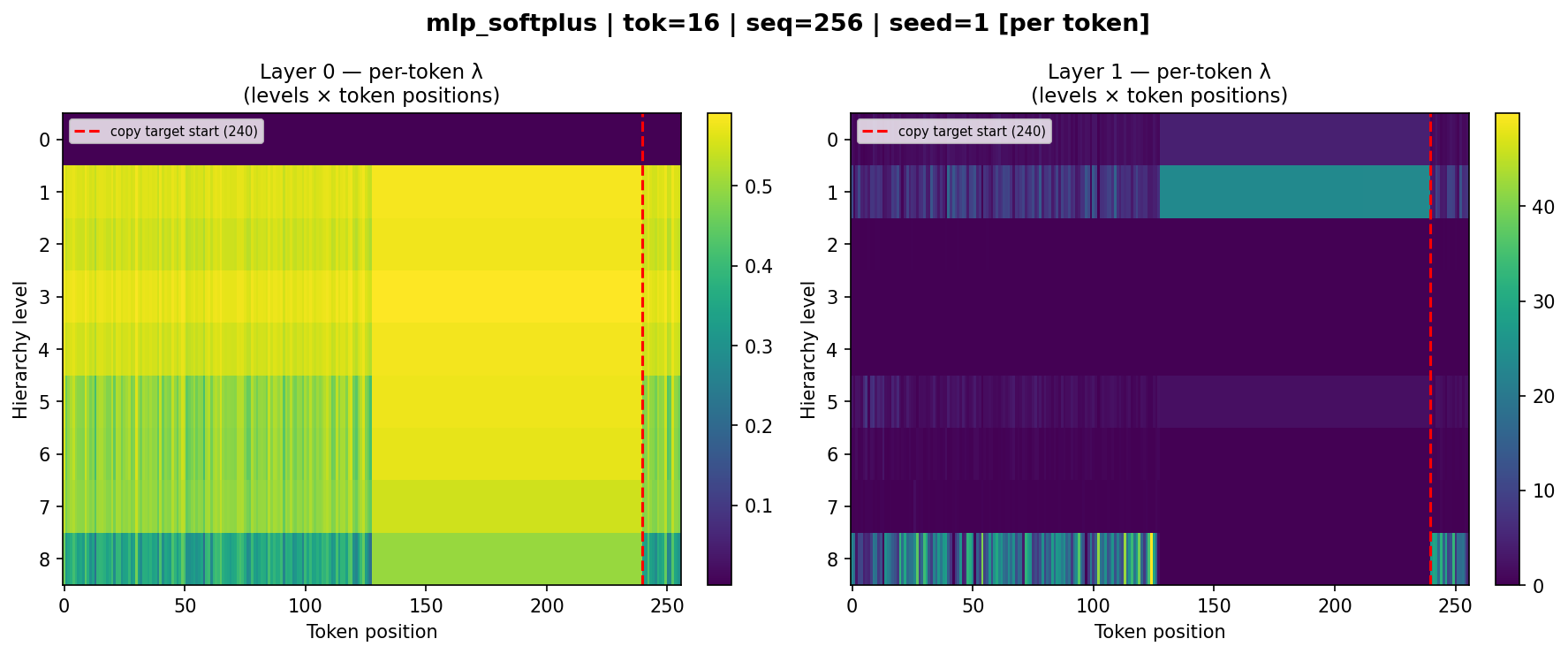}
\end{minipage}
\vspace{0.2em}
\begin{minipage}{0.49\textwidth}
    \centering
    \includegraphics[width=\linewidth]{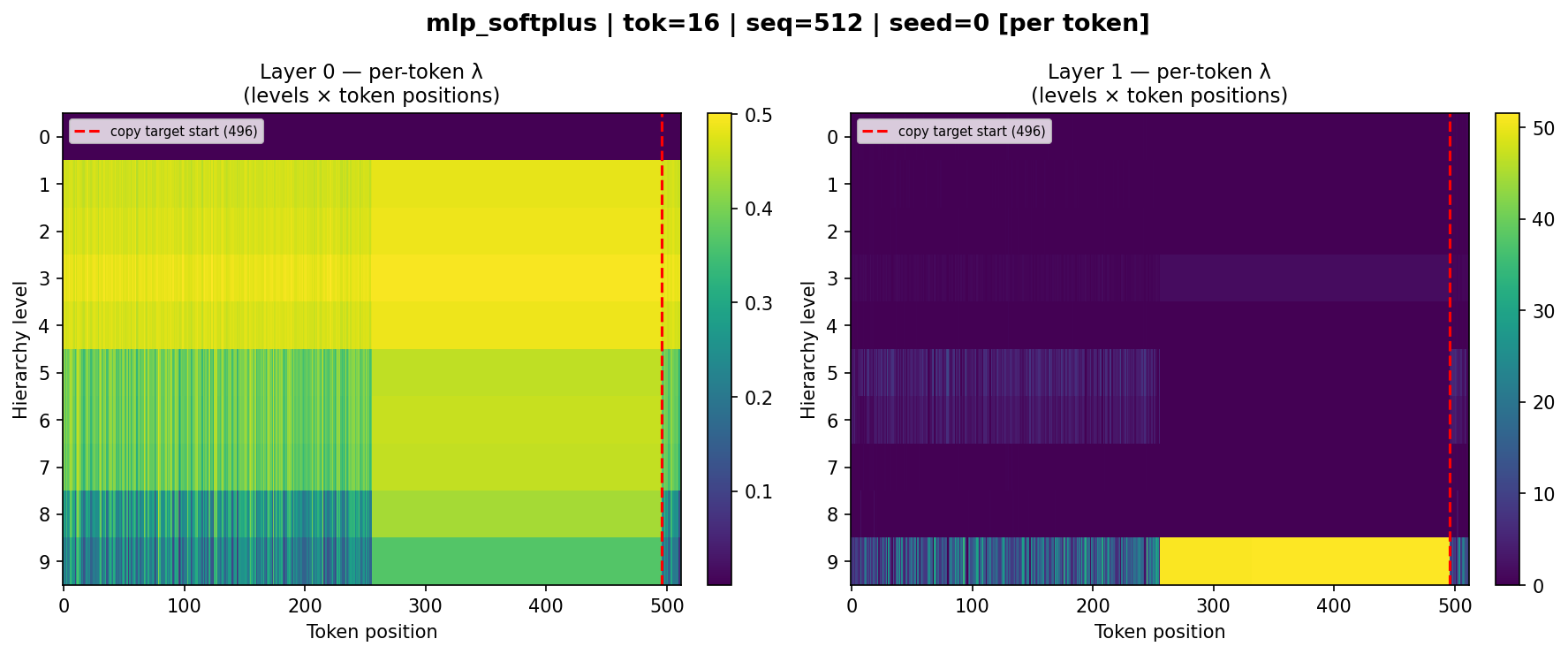}
\end{minipage}
\hfill
\begin{minipage}{0.49\textwidth}
    \centering
    \includegraphics[width=\linewidth]{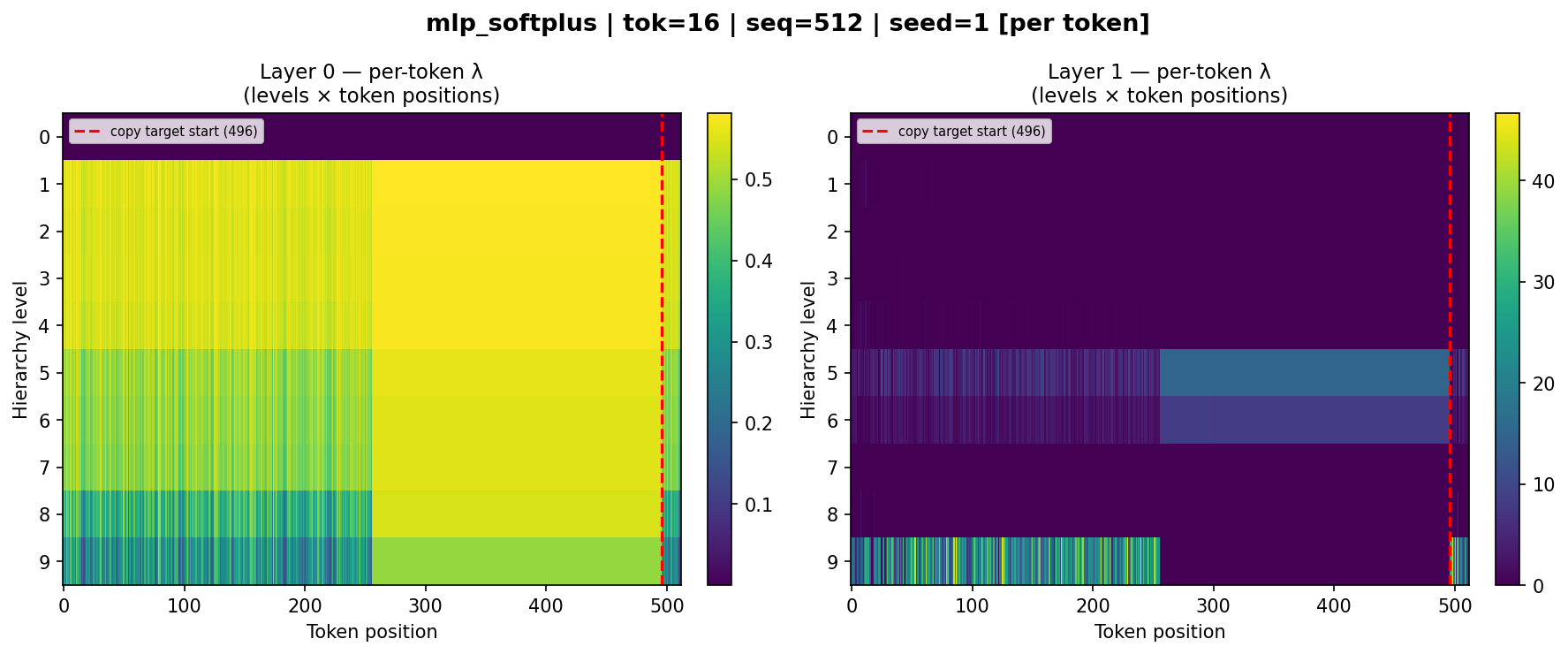}
\end{minipage}
\vspace{0.2em}
\begin{minipage}{0.49\textwidth}
    \centering
    \includegraphics[width=\linewidth]{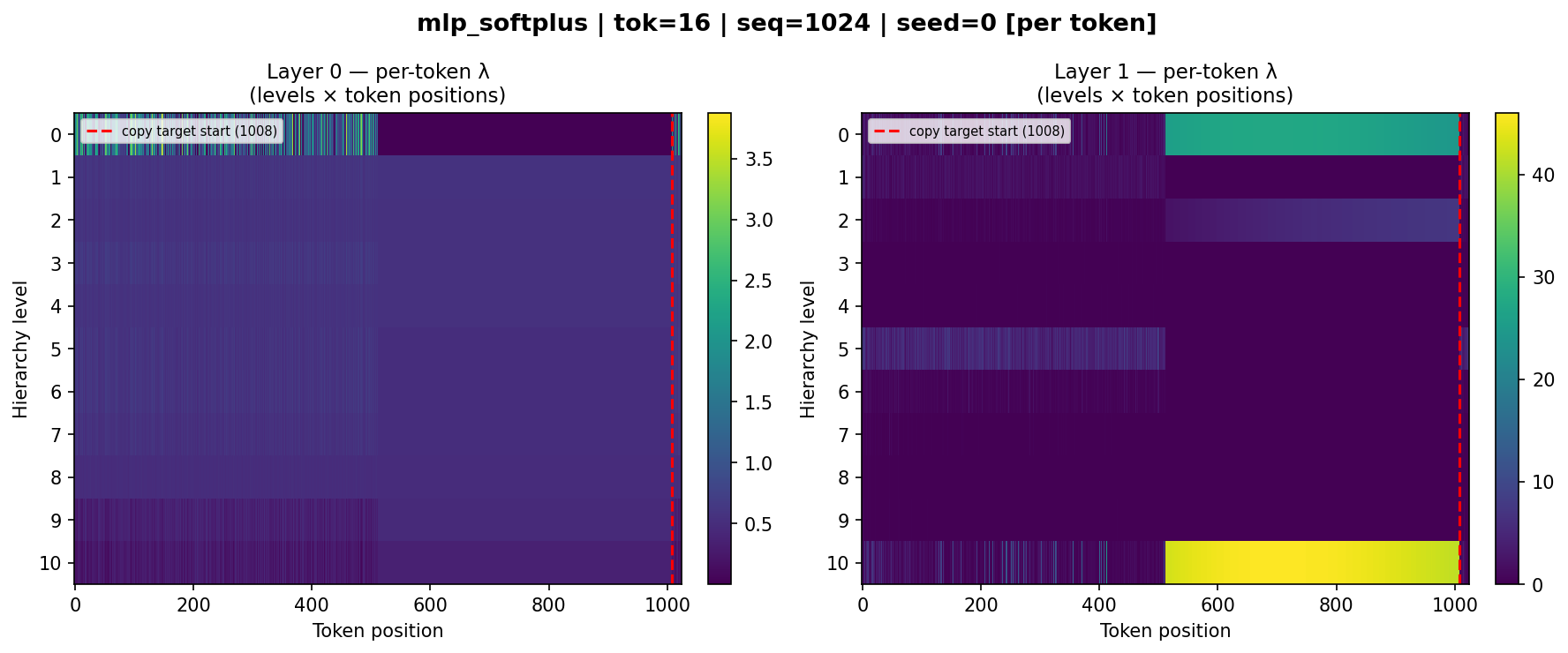}
\end{minipage}
\hfill
\begin{minipage}{0.49\textwidth}
    \centering
    \includegraphics[width=\linewidth]{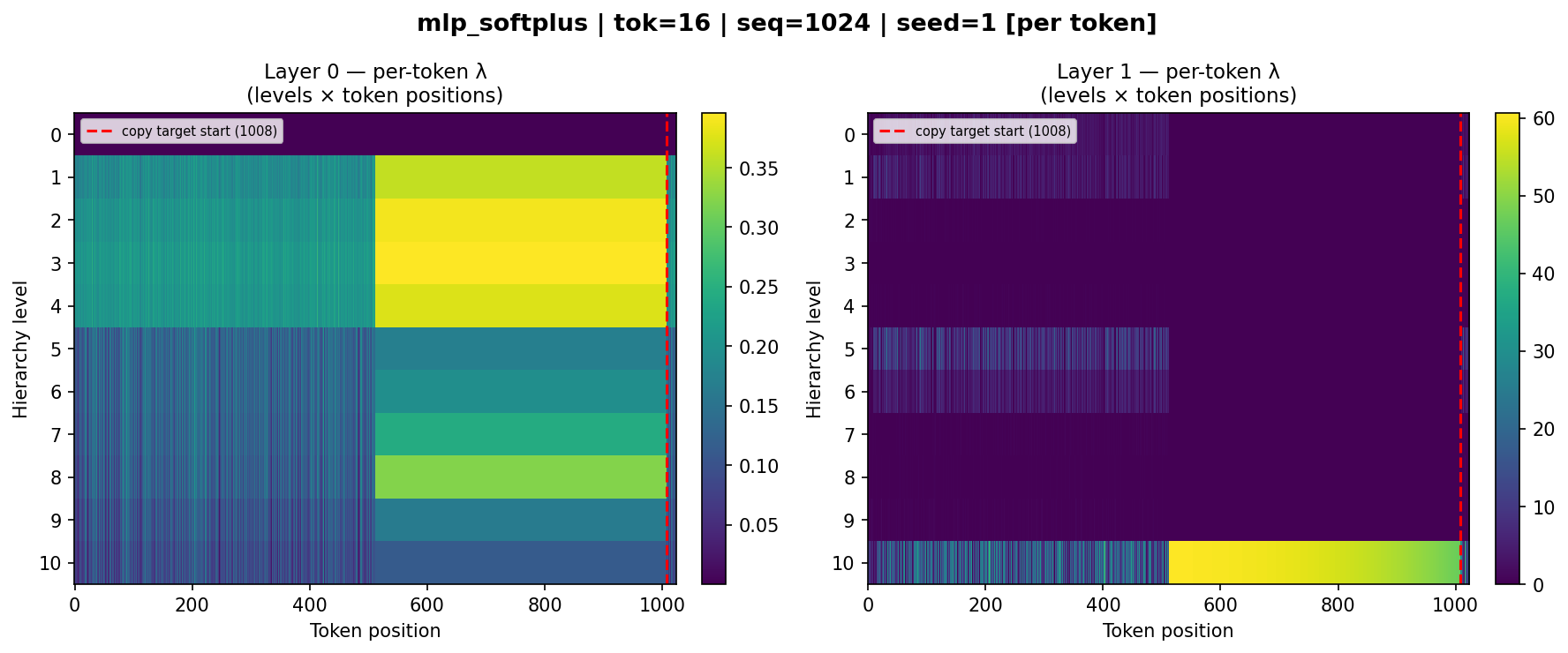}
\end{minipage}
\caption{Token-level MLP-softplus $\lambda$ heatmaps on selective copying for
tok=16 across sequence lengths 256, 512, and 1024. The MLP-softplus
parameterization produces token-dependent memory-level weights, with visible
changes near the copy target region marked by the red dashed line.}
\label{fig:sc_softplus_all}
\end{figure}

Figure~\ref{fig:sc_softplus_all} shows the corresponding token-level heatmaps
for MLP-softplus. Unlike the baseline, the MLP-softplus weights vary
across both token positions and hierarchy levels. The learned patterns also
change near the later part of the sequence, where copying becomes relevant.
This suggests that the MLP-parameterized $\lambda$ can adjust memory-level
weighting based on the current position's role in the task, rather than using
one global layer-wise profile for the whole sequence. This behavior remains
visible at sequence length 1024, where the task places more pressure on
long-range memory.

\begin{figure}[htbp]
\centering
\begin{minipage}{0.49\textwidth}
    \centering
    \includegraphics[width=\linewidth]{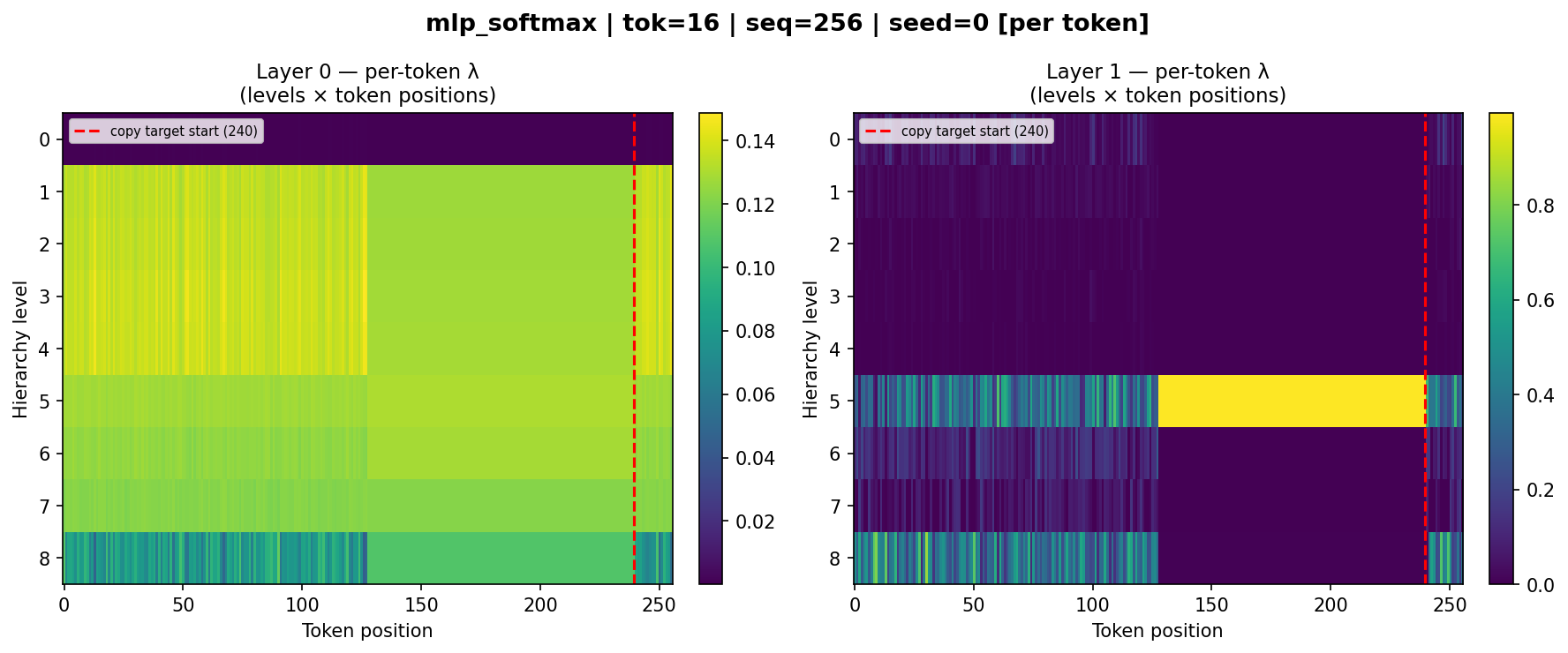}
\end{minipage}
\hfill
\begin{minipage}{0.49\textwidth}
    \centering
    \includegraphics[width=\linewidth]{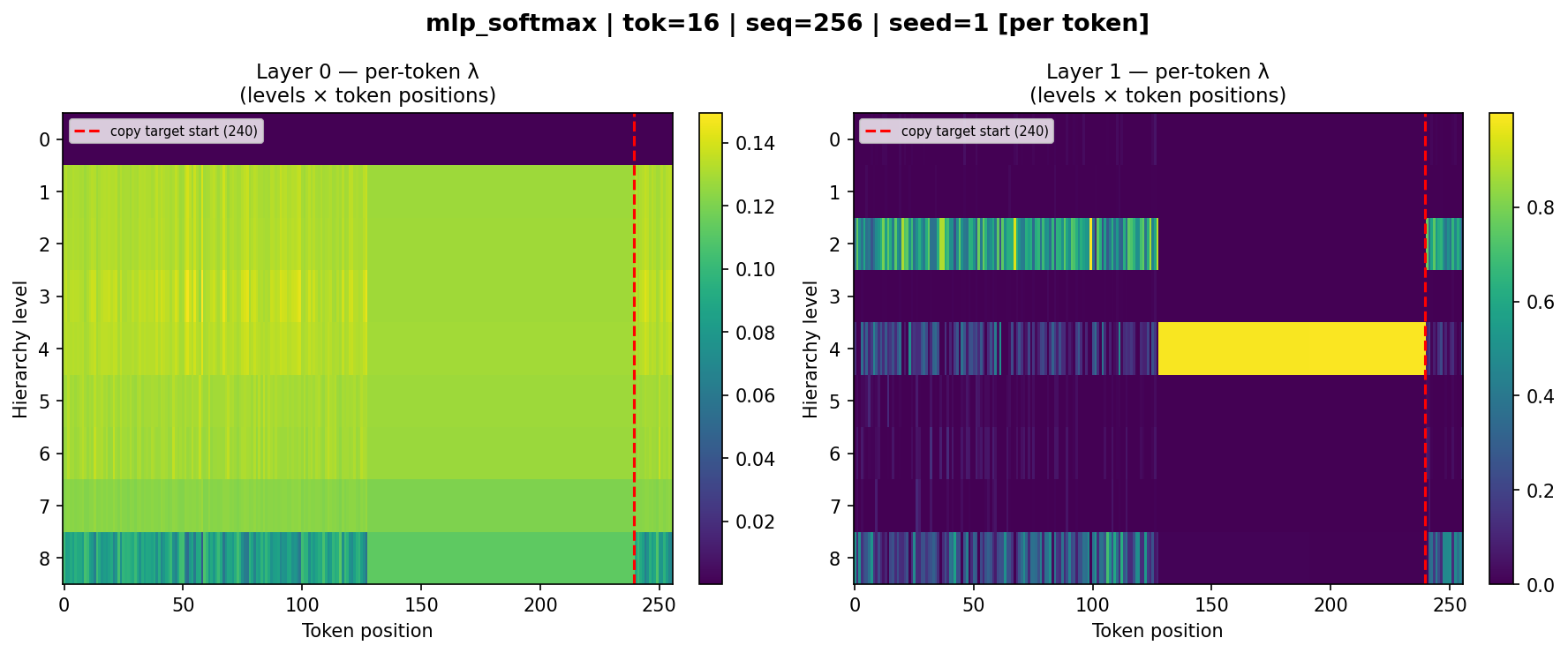}
\end{minipage}
\vspace{0.2em}
\begin{minipage}{0.49\textwidth}
    \centering
    \includegraphics[width=\linewidth]{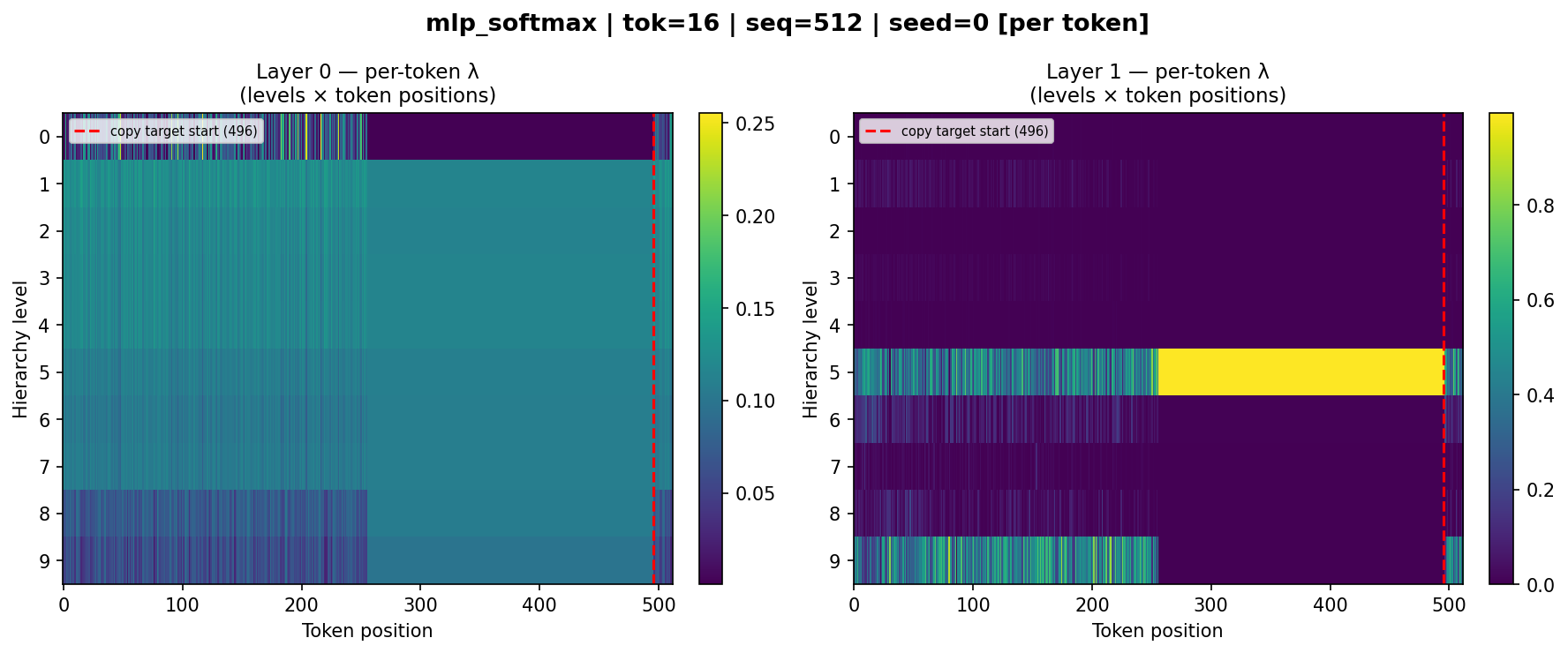}
\end{minipage}
\hfill
\begin{minipage}{0.49\textwidth}
    \centering
    \includegraphics[width=\linewidth]{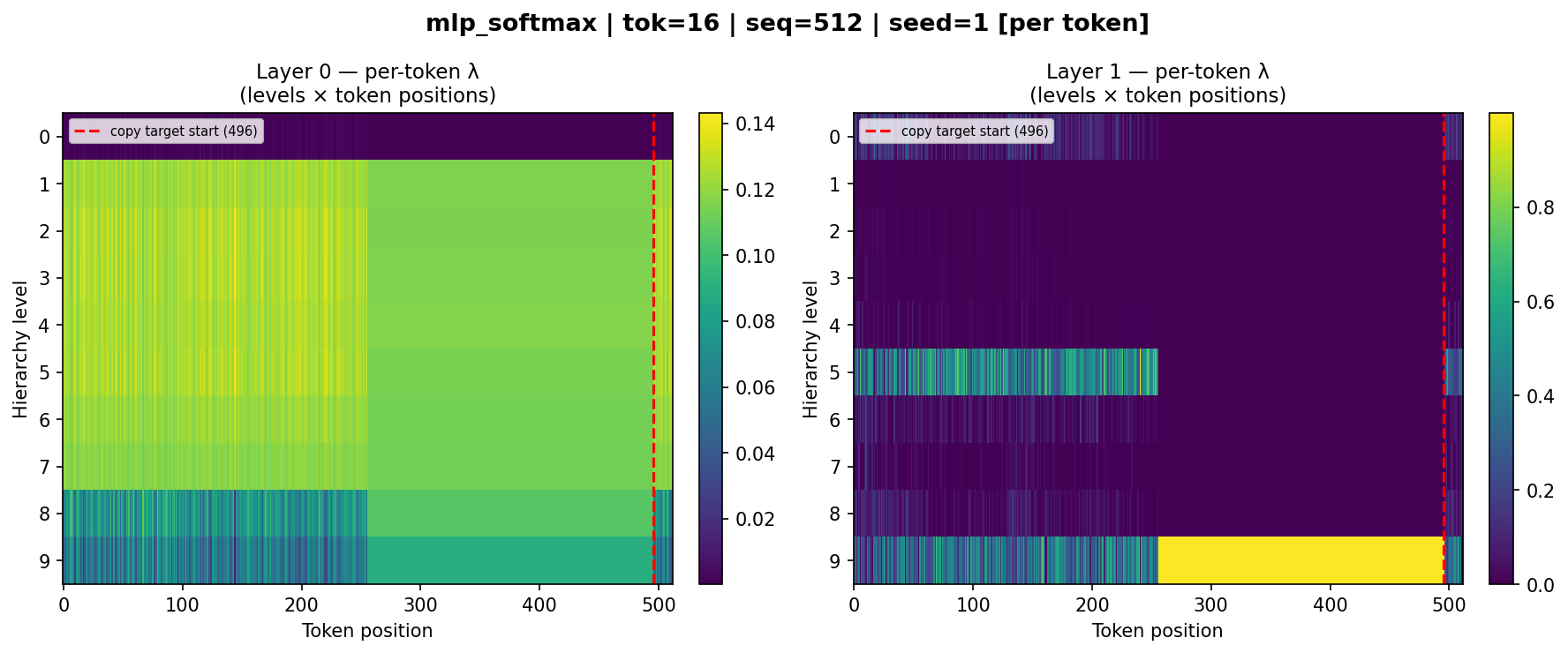}
\end{minipage}
\vspace{0.2em}
\begin{minipage}{0.49\textwidth}
    \centering
    \includegraphics[width=\linewidth]{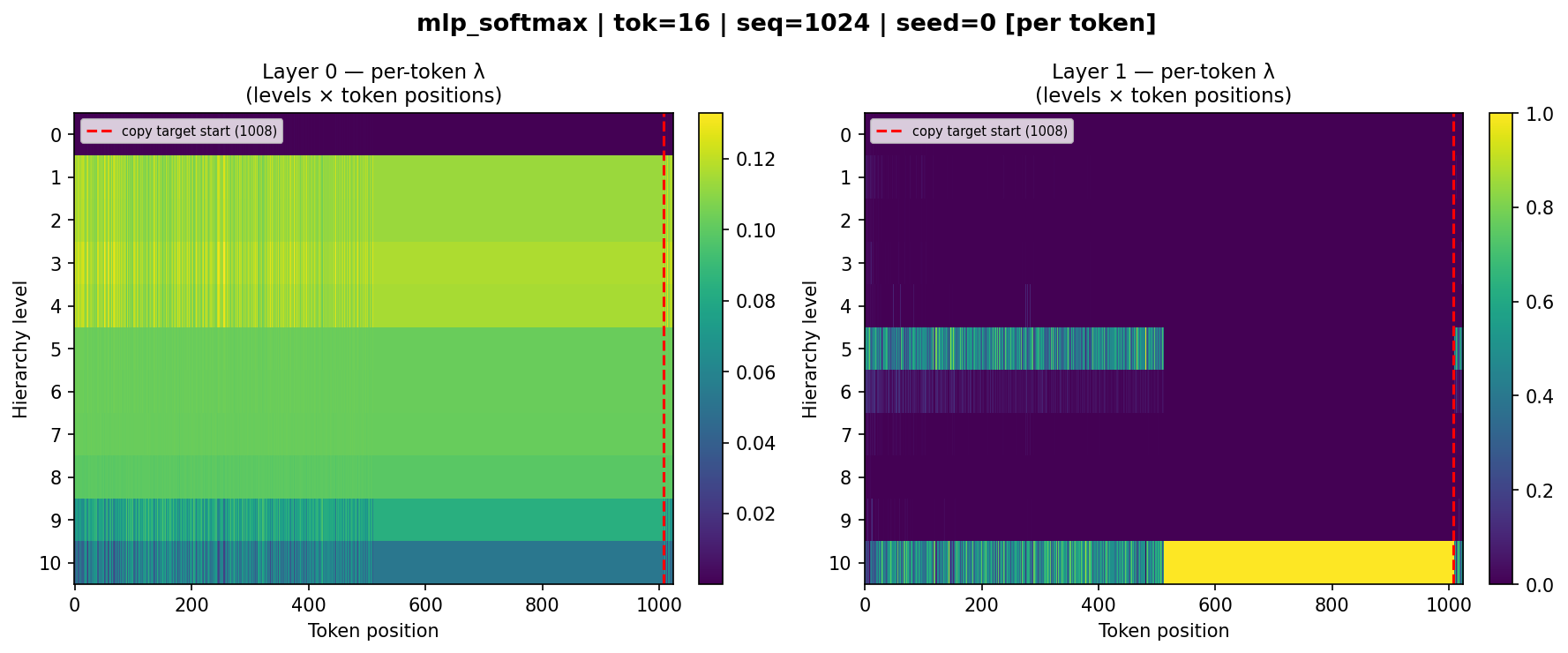}
\end{minipage}
\hfill
\begin{minipage}{0.49\textwidth}
    \centering
    \includegraphics[width=\linewidth]{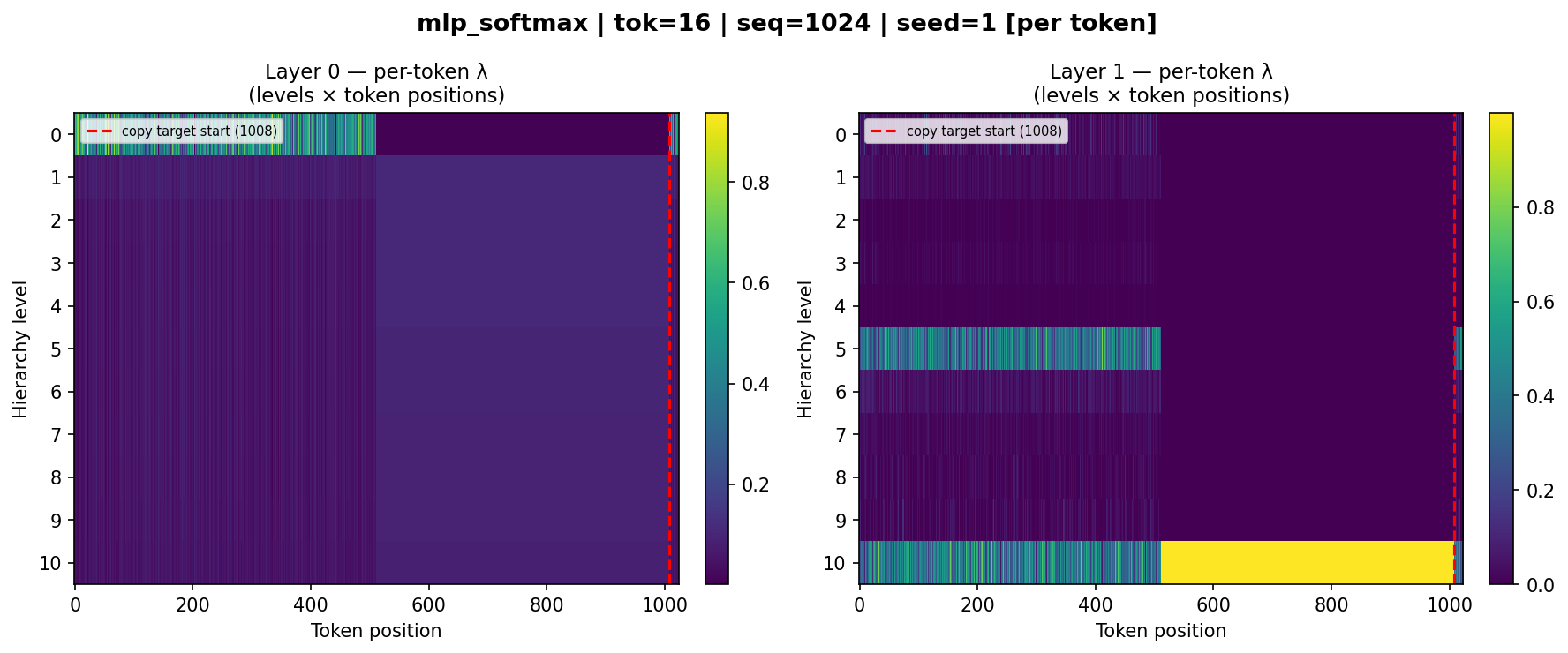}
\end{minipage}
\caption{Token-level MLP-softmax $\lambda$ heatmaps on selective copying for
tok=16 across sequence lengths 256, 512, and 1024. MLP-softmax also produces
token-dependent memory-level weights, but its normalization across levels
often leads to sharper and more seed-sensitive profiles.}
\label{fig:sc_softmax_all}
\end{figure}

For completeness, Figure~\ref{fig:sc_softmax_all} shows the MLP-softmax
variant on the same task. As in the MQAR visualizations, softmax normalization
encourages sharper competition across hierarchy levels. The resulting profiles
remain token-dependent, but the active levels can be more concentrated and more
sensitive to the seed than in the MLP-softplus variant. This provides a useful
comparison between the two adaptive parameterizations: both allow token-level
variation, but softplus tends to produce smoother memory weighting while
softmax produces more competitive level selection.

\begin{figure}[htbp]
\centering
\begin{minipage}{0.32\textwidth}
    \centering
    \includegraphics[width=\linewidth]{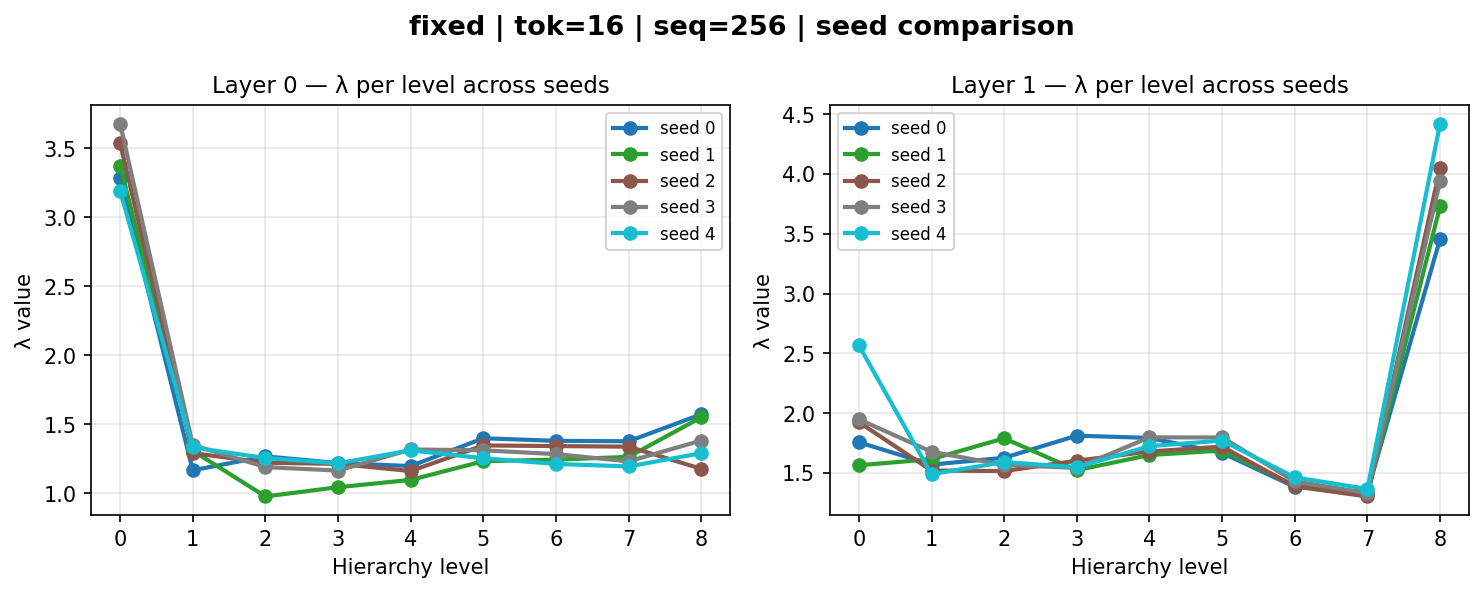}
\end{minipage}
\hfill
\begin{minipage}{0.32\textwidth}
    \centering
    \includegraphics[width=\linewidth]{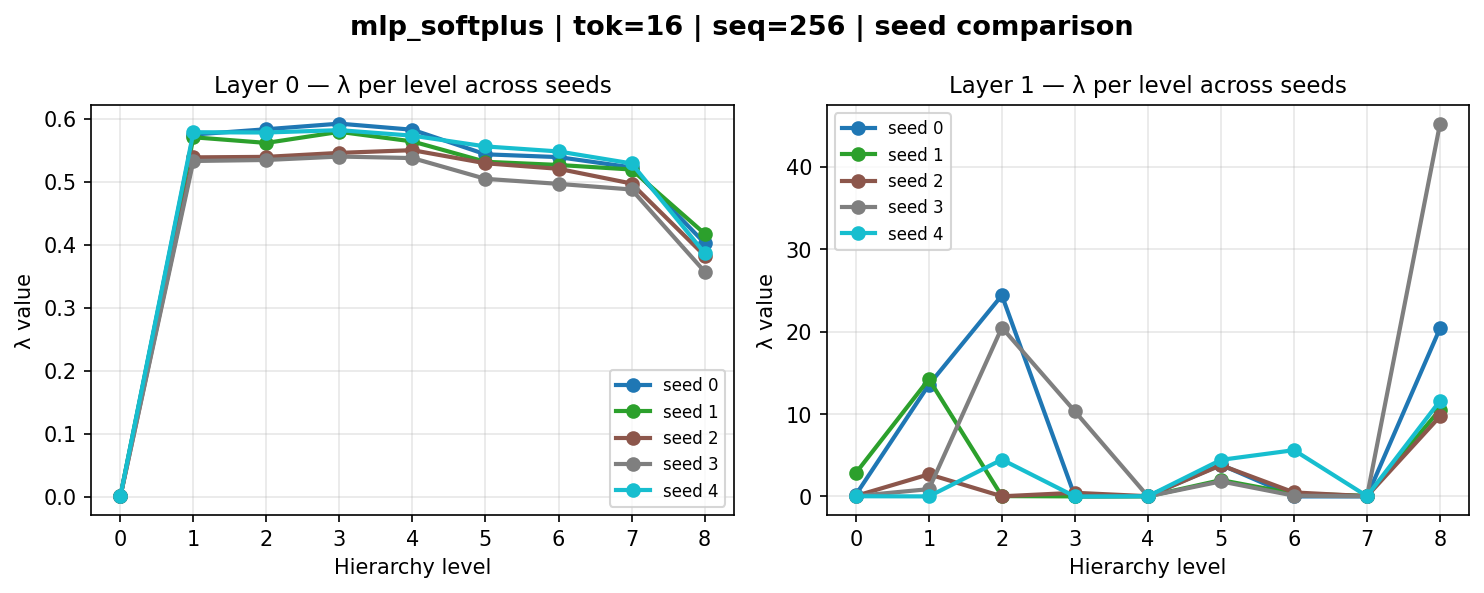}
\end{minipage}
\hfill
\begin{minipage}{0.32\textwidth}
    \centering
    \includegraphics[width=\linewidth]{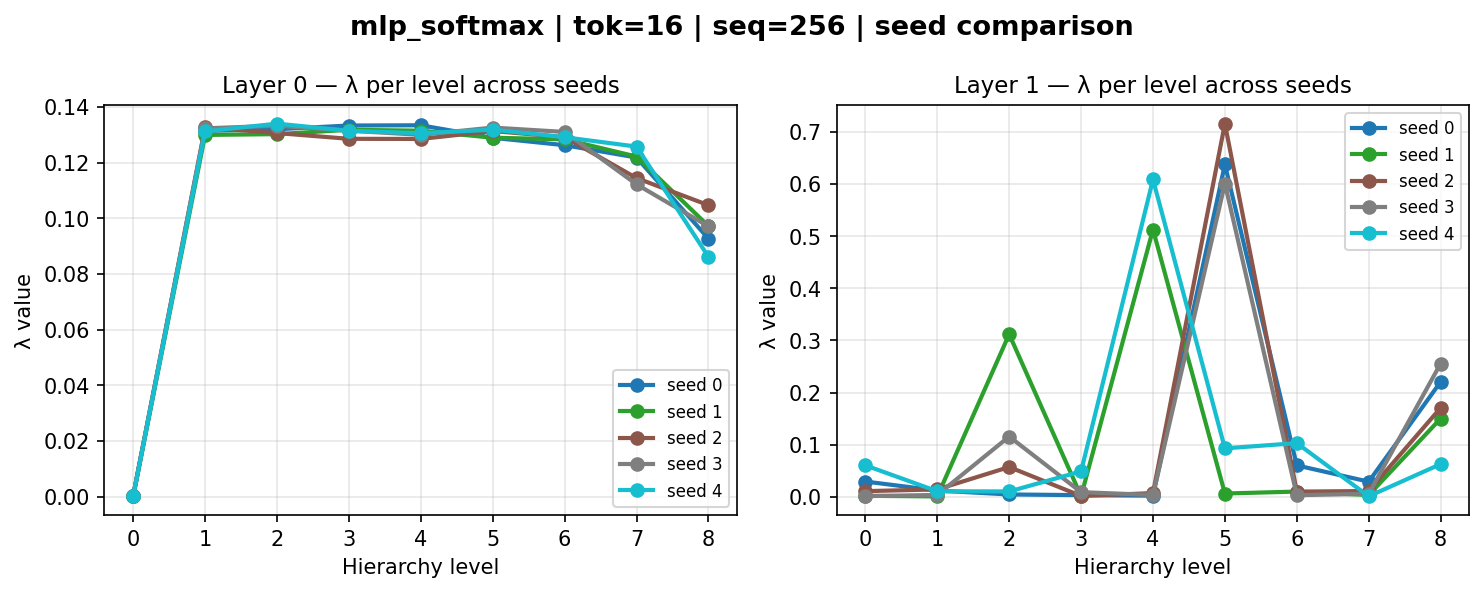}
\end{minipage}
\caption{Seed comparison of learned $\lambda$ profiles on selective copying
for tok=16 and sequence length 256. The baseline learns stable but
static layer-wise profiles, while MLP-softplus and MLP-softmax produce more
specialized profiles with stronger variation across seeds and layers.}
\label{fig:sc_seed_comp}
\end{figure}

Finally, Figure~\ref{fig:sc_seed_comp} summarizes the learned $\lambda$
profiles across seeds for selective copying at sequence length 256. The fixed
baseline is stable but static, while MLP-softplus and MLP-softmax produce more
specialized level profiles. This mirrors the MQAR seed comparisons and
supports the same qualitative conclusion across tasks: fixed $\lambda$ learns
one global memory strategy per layer, whereas adaptive $\lambda$
parameterizations allow the model to change how it uses the hierarchy.

\end{document}